\documentclass[10pt,twocolumn,letterpaper]{article}

\usepackage[pagenumbers]{cvpr} %

\usepackage{tikz}
\usepackage{multirow}
\usepackage{makecell}
\usepackage{algorithm}
\usepackage[noend]{algorithmic}
\usepackage{amsmath, amsthm}

\newtheorem{theorem}{Theorem}
\newtheorem{lemma}{Lemma}
\newtheorem{assumption}{Assumption}
\newcommand{\restate}[2]{\noindent\textbf{Theorem \ref{#1} (restated).} \textit{#2}}

\def\maketitlesupplementaryonecolumn
   {
   \newpage
       \onecolumn
       {
        \centering
        \Large
        \textbf{\thetitle}\\
        \vspace{0.5em}Supplementary Material \\
        \vspace{1.0em}
        }
   }

\providecommand{\argmax}{\mathop\mathrm{arg\, max}} 

\usepackage[most]{tcolorbox}
\usepackage{multirow}
\definecolor{cvprblue}{rgb}{0.21,0.49,0.74}
\usepackage[pagebackref,breaklinks,colorlinks,allcolors=cvprblue]{hyperref}

\def\paperID{16965} %
\def\confName{CVPR}
\def\confYear{2025}

\title{SPARC: Score Prompting and Adaptive Fusion for Zero-Shot Multi-Label Recognition in Vision-Language Models}

\author{
Kevin Miller\\
Boston University\\
{\tt\small nivek@bu.edu}
\and
Aditya Gangrade\\
Boston University\\
{\tt\small gangrade@bu.edu}
\and
Samarth Mishra\\
Boston University\\
{\tt\small samarthm@bu.edu}
\and
Kate Saenko\\
Boston University and Meta AI (FAIR)\\
{\tt\small saenko@bu.edu}
\and
Venkatesh Saligrama\\
Boston University\\
{\tt\small srv@bu.edu}
}

\begin{document}
\maketitle
\newcommand{\captionMainTableCOCO}{Main results for COCO.}
\newcommand{\captionMainTableVOC}{Main results for VOC.}
\newcommand{\captionMainTableNUSWIDE}{Main results for NUSWIDE.}
\newcommand{\captionMainTableAvgArch}{Results over three datasets, averaged over backbones.}
\newcommand{\captionCompetitionTagCLIPlogNOrowcalibbaseNO}{Results from combining our method with TagCLIP.}
\newcommand{\captionCompetitionTagCLIPlogNOrowcalibbaseYES}{Results from combining our method with TagCLIP.}
\newcommand{\captionCompetitionTagCLIPlogYESrowcalibbaseNO}{Shows complementarity with architectural approaches, improving TagCLIP by 2.6\%}
\newcommand{\captionCompetitionTagCLIPlogYESrowcalibbaseYES}{Results from combining our method with TagCLIP.}
\newcommand{\captionCompetitionTaIDPTrowcalibbaseNO}{TaI-DPT results}
\newcommand{\captionCompetitionTaIDPTrowcalibbaseYES}{TaI-DPT results}
\newcommand{\captionCompetitionCoMCrowcalibbaseNO}{CoMC results}
\newcommand{\captionCompetitionCoMCrowcalibbaseYES}{CoMC results}
\newcommand{\captionComphrehensiveCompoundOnly}{Average mAP (across all datasets and architectures) for different Rank Fusion strategies, without ``merge'' step.}
\newcommand{\captionComphrehensiveAvg}{Average mAP (across all datasets and architectures) for different Rank Fusion strategies, with ``merge'' step.}
\newcommand{\captionComphrehensivePCA}{Average mAP (across all datasets and architectures) for different selection strategies, PCA ensemble.}
\newcommand{\captionNoiseModelResults}{Comparing different noise models for pair prompt scores.}
\newcommand{\captionDebiasAblationTable}{Ablations on Debias module. Quantifies impact of debiasing both with and without compound prompts.}
\newcommand{\captionQualitativeHist}{Histograms for singleton, 1st max, and 2nd max scores for ``cat'' in the COCO dataset. We see that 1st max creates overlap by lifting the scores of some ground-truth negatives. 2nd max does not create these issues and performs well when fused with singleton scores.}
\newcommand{\captionQualitativeExample}{An example of why 2max works better than 1max. The left image gets a lower singleton score than the right image, resulting in a misordering. The 1max strategy lifts the scores of both images and thus fails to correct the misordering. The 2max strategy successfully corrects the misordering by lifting only the left image's score.}
\newcommand{\captionOurMethodSchematic}{A schematic of our method. \textbf{Top:} We use the target classnames along with cooccurrence stats to create compound prompts that mention multiple classes together. \textbf{Bottom:} At inference time, we query the VLM with both singleton and compound prompts, and use debiasing, selection, and fusion to come up with refined scores.}
\newcommand{\captionIntroFigure}{\textbf{Top Left:} Motivating example where the second-highest-scoring compound prompt does a better job than the highest-scoring one at correcting a misordering. \textbf{Top Right:} Our method uses classnames and cooccurrence info to generate compound prompts that mention multiple objects. \textbf{Bottom Left:} We use Debias and Rank Fusion modules to refine singleton scores by combining them with compound scores. \textbf{Bottom Right:} Rank Fusion module sorts the compound scores weights them based on max variation direction.}
\newcommand{\captionMethodFigure}{(top) Vision Language models(VLMs) like CLIP can be used for zero-shot classification with image-text similarity scores. While this works fairly well for single-class labels, they can struggle in the multi-label scenario. (bottom) In this paper, we instroduce SPARC, our solution that functions on top of an existing VLM, treating it simply as a black-box score generator. Using class names, SPARC first creates compound prompts for additional queries to the VLM. It then debiases, ranks and appropriately fuses them to generate final scores for the original classes.}
\newcommand{\captionExampleFigure}{A motivating example with an image where class ``cat'' is absent (left) and one where it is present (right). The highest compound prompt score is an unhelpful signal because it gives a high score to both negatives and positives, while the second-highest is more discriminative. Our method adaptively fuses the most informative order statistics, resulting in a strong signal.}
\newcommand{\captionCompetitionTaIDPTandCoMC}{Results from combining our method with TaI-DPT (left) and CoMC (right) showing compatibility with training-based methods. Our method improves TaI-DPT by 1.7\% while preserving the existing strong signal of CoMC, degrading it by only 0.3\%.}%
\newcommand{\captionMainTableThreeDatasets}{Results over three datasets and nine CLIP backbones. Our proposed method consistently outperforms the ZSCLIP baseline across all datasets and architectures exhibiting its effectivness on multi-label recognition tasks.}
\newcommand{\captionRankFusionAblation}{Average mAP for different Rank Fusion strategies, without (top) and with (bottom) the ``merge'' step \eqref{eq:merge} demonstrates superiority of adaptive fusion over fixed strategies.}
\newcommand{\captionPerClassAPsCOCO}{Per-class APs (averaged over all CLIP backbones) for our method vs vanilla ZSCLIP on the COCO dataset. Our method consistently improves over ZSCLIP for almost every class. Plots for VOC and NUSWIDE are shown in the Supplementary.}

\begin{abstract}
Zero-shot multi-label recognition (MLR) with Vision-Language Models (VLMs) faces significant challenges without training data, model tuning, or architectural modifications. Existing approaches require prompt tuning or architectural adaptations, limiting zero-shot applicability. Our work proposes a novel solution treating VLMs as black boxes, leveraging scores without training data or ground truth. Using large language model insights on object co-occurrence, we introduce compound prompts grounded in realistic object combinations. Analysis of these prompt scores reveals VLM biases and ``AND''/``OR'' signal ambiguities, notably that maximum compound scores are surprisingly suboptimal compared to second-highest scores. We address these through a debiasing and score-fusion algorithm that corrects image bias and clarifies VLM response behaviors. Our method enhances other zero-shot approaches, consistently improving their results. Experiments show superior mean Average Precision (mAP) compared to methods requiring training data, achieved through refined object ranking for robust zero-shot MLR.
\end{abstract}
    
\vspace{-12pt}
\section{Introduction}
\begin{figure}[t]
  \centering
   \includegraphics[width=1\linewidth]{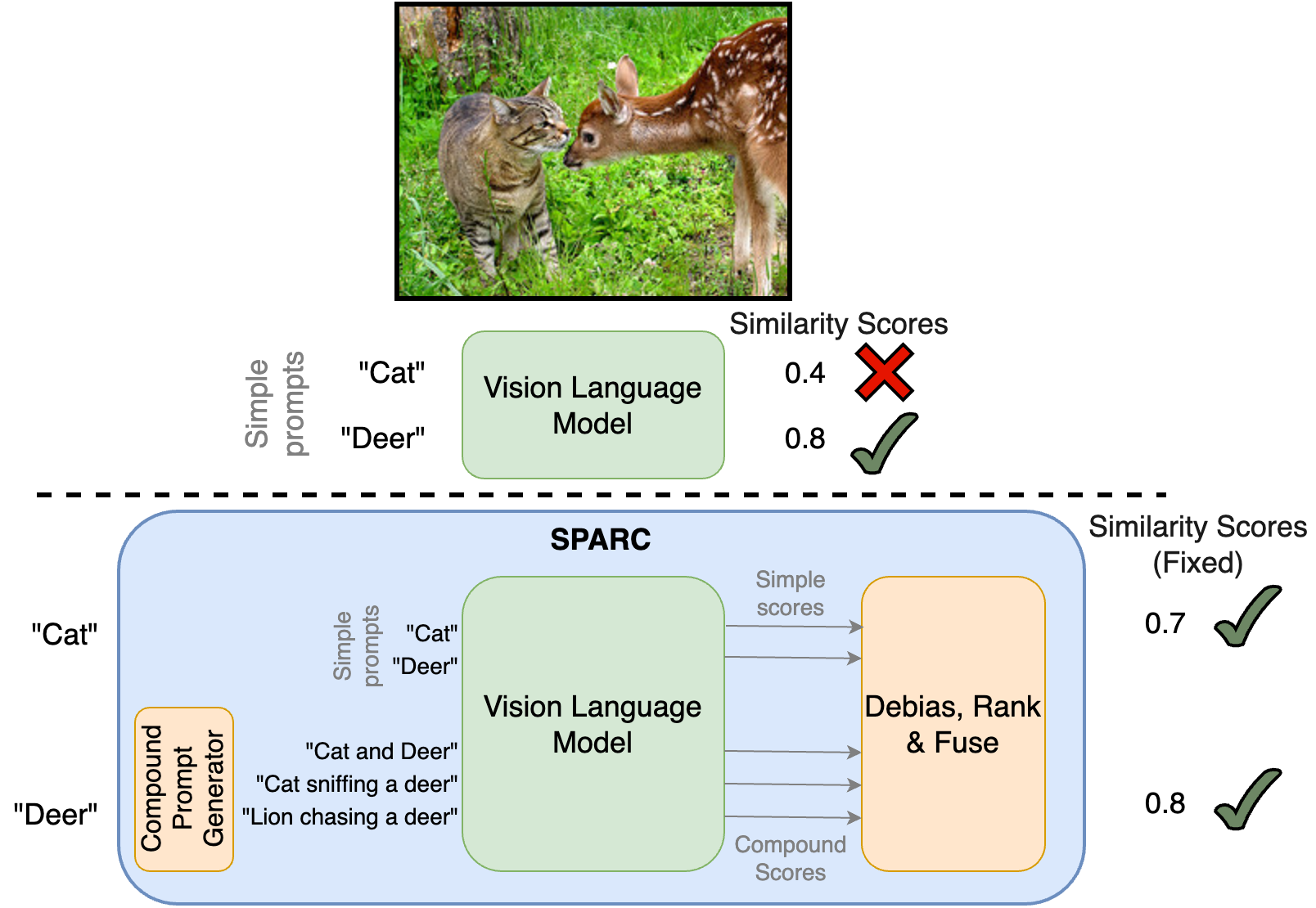}
   \caption{\captionMethodFigure}
   \label{fig:methodFigure}
\end{figure}
\begin{figure}[t]
  \centering
   \includegraphics[width=1\linewidth]{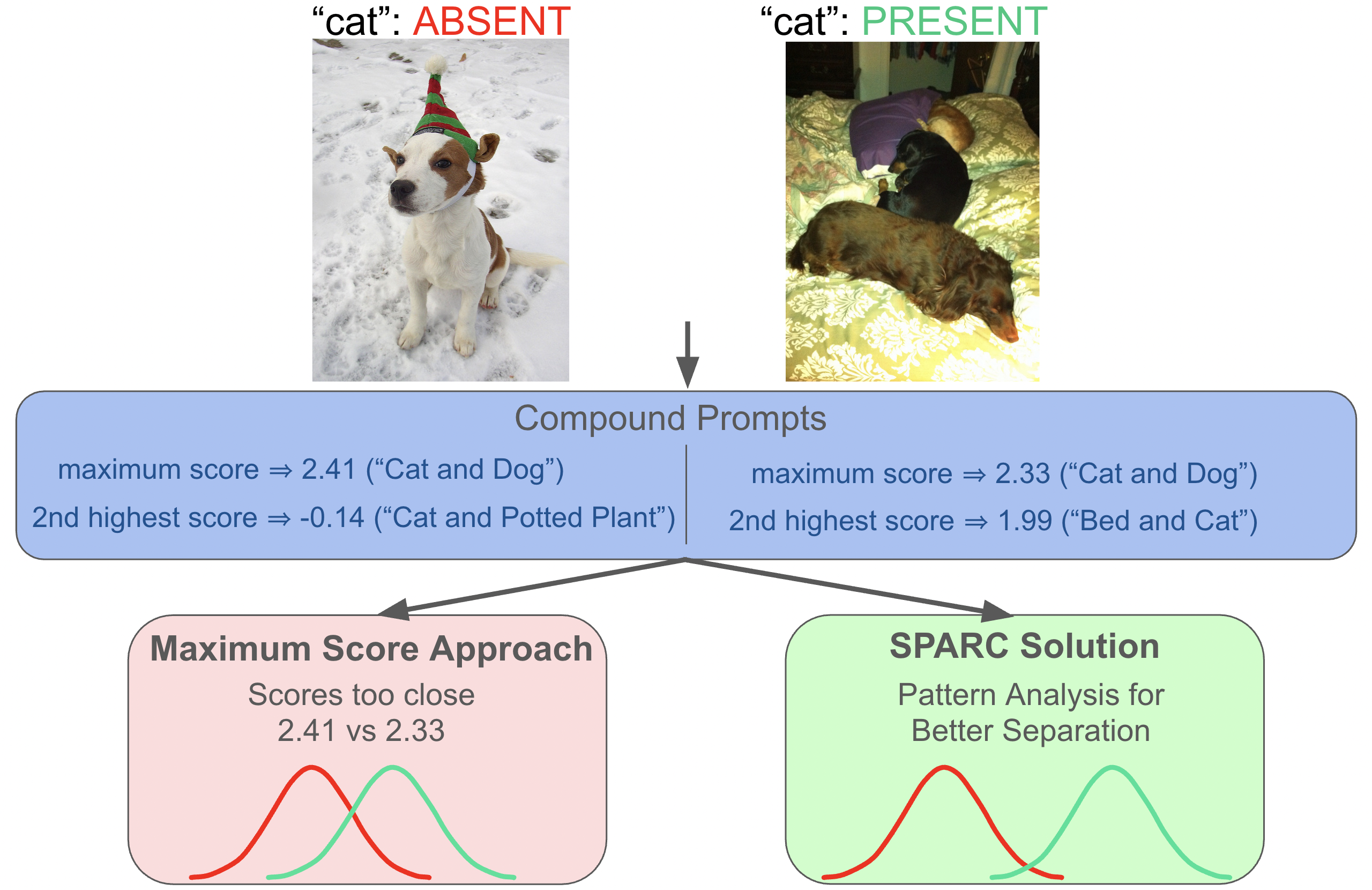}
   \caption{\captionExampleFigure}
   \label{fig:exampleFigure}
   \vspace{-12pt}
\end{figure}
Vision-Language Models (VLMs), such as CLIP~\cite{CLIP}, have emerged as general-purpose systems for understanding visual data through language-based queries. These models enable a broad range of applications, from object detection to image captioning, by linking visual inputs to language prompts. In standard settings where images contain single, recognizable objects, VLMs perform remarkably well. However, for the more complex task of zero-shot multi-label recognition (MLR) (Fig. \ref{fig:methodFigure} (top)), where models must identify multiple objects within an image without prior training on specific data, VLMs face significant limitations. Zero-shot MLR is crucial for applications in fields like robotics and medical imaging, where objects rarely appear in configurations that align neatly with training distributions. In these scenarios, achieving robust multi-label recognition without fine-tuning is challenging, given the task’s reliance on mean Average Precision (mAP) scores, which depend on ranking images for object presence.

\noindent \textbf{VLM: Prompt Dependent AND/OR Noisy Channel.} Despite the promise of zero-shot capabilities, current VLM approaches often struggle with MLR due to inherent scoring behaviors and biases. The performance of these models is hampered by a mix of conjunction (AND) and disjunction (OR) behaviors in their scoring, leading to inflated scores in compound prompts that contain multiple objects. For example, a prompt like “cat and sofa” might yield a high score even if only one of these objects is present in the image. This tendency reflects biases learned during training, where common object pairs receive higher scores even when only one object is present, disrupting the accuracy of mAP-based evaluations. Furthermore, existing methods for adapting VLMs to zero-shot MLR frequently rely on prompt tuning or architectural adjustments—approaches that are often dependent on training data and computationally intensive fine-tuning, which limit their generalizability to novel tasks.

\noindent \textbf{Our Approach.} In contrast to these methods, we introduce SPARC (Score Prompting and Adaptive Fusion for Zero-Shot Multi-Label Recognition in VLMs), a novel approach to zero-shot MLR that bypasses the need for training data, prompt tuning, or model-specific modifications. Our method treats the VLM as a black box, relying solely on its output scores to infer object presence (see Fig. \ref{fig:methodFigure}). This black-box approach enables us to avoid assumptions about the model’s internal workings, allowing for a purely zero-shot framework that is both model-agnostic and dataset-independent. SPARC introduces two main innovations that address the unique challenges of zero-shot MLR.

\noindent \textbf{A. Compound Prompt Composition:} Recognizing that VLMs can provide richer information when prompted with combinations of objects, we develop a method for constructing compound prompts. These prompts reflect likely contextual associations between objects, such as “cat and sofa” or “car and bus.” By gathering scores from these compound prompts, we can capture a spectrum of potential object contexts within the image, enhancing detection without relying on training-based adaptations. This composition strategy allows us to agnostically extract information from the VLM, leveraging probable object relationships without depending on any specific dataset or VLM architecture.\\
\noindent \textbf{B. Score Debiasing and Adaptive Fusion.} A critical insight in our approach lies in the surprising observation that the \underline{maximum score among compound prompts} is often a poor proxy for true object presence. Although one might expect the highest score to serve as a reliable signal, we find that it frequently reflects compositional biases, as VLMs tend to respond to compound prompts with OR-like behavior, raising scores even when only one object in the prompt is present. Instead, we observe that the second-highest score consistently provides a more accurate indicator of object presence, minimizing the effects of false positives caused by compositional bias. Building on this insight, we develop a debiasing algorithm that normalizes scores across images to address image-specific noise and clarify genuine object presence signals. This debiased score set is then processed through a PCA-based fusion method that further refines object rankings by combining information from both compound and singleton prompts, ultimately optimizing mAP by enhancing score accuracy.\\
\textbf{Complementarity.} SPARC is complementary to other zero-shot and training-free MLR methods. When applied on top of these approaches, SPARC consistently enhances mAP scores by refining object ranking and reducing bias in VLM outputs. This capability makes SPARC an adaptable solution that can improve upon existing methods while maintaining a fully zero-shot, model-agnostic framework.

\noindent \textbf{Empirical Results.} SPARC achieves significant improvements in mAP, outperforming methods that incorporate architectural modifications. This outcome shows the potential of a fully zero-shot approach that relies only on systematic prompt design and score interpretation, rather than prompt-training or fine-tuning. By revealing that the second-highest score can be a superior proxy to the maximum, our findings provide new insights into VLM scoring behavior, suggesting that careful treatment of prompt compositions and score patterns can unlock robust MLR capabilities.

\subsection{Related Work}
\label{sec:related}

\noindent \textbf{Supervised Methods.} 
Prior methods improve MLR performance through various approaches. DualCoOp \cite{DualCoOp,DualCoOp++} trains prompts for text-guided spatial attention. Subsequent works use class co-occurrences to refine CLIP logits \cite{CoocGCN,HSPNet}. Hierarchical structuring \cite{HSPNet} encourages related classes to learn similar prompts. SSPA \cite{SSPA} combines embeddings from learned and LLM-generated prompts, and TRM-ML \cite{TRM-ML} uses pseudolabels to match texts to image regions. While effective, these methods require annotated training data. In contrast, ours is unsupervised.

\noindent \textbf{Unsupervised Training-Based Methods.} 
TaI-DPT \cite{TaI-DPT} uses caption-only data to adapt CLIP to MLR by training a prompt-generation network using caption embeddings. Extensions include LLM-generated captions \cite{DataFreeMLR}, pseudo-visual prompts \cite{TaiPlusPlus}, and lightweight classifiers on caption embeddings \cite{CoMC, TaI-Adapter}. Separately, CDUL \cite{CDUL} leverages CLIP scores with spatial aggregation to create training pseudolabels. While these approaches demonstrate shared vision-language embedding power, they require significant training and CLIP internal access. In contrast, our black-box method uses crude cooccurrence-based compound prompts (essentially deleting implausible contextual associations) to probe multilabel information, fusing scores for effective MLR.

\noindent\textbf{Unsupervised Training-free methods.} The recent literature has proposed methods to adapt CLIP to the MLR taks in a training-free manner via architectural changes. CLIPSurgery \cite{CLIPSurgery} alters the architecture to increase the coherence of attention masks with in the image backbone, and debiases features on a patch-by-patch basis. TagCLIP \cite{TagCLIP} uses masks from within the CLIP image backbone to refine scores at a patch level based on regional coherence, and aggregate queries over crops of the image. Again, both of these methods require white-box access to the internals of the VLM model, while ours works in a black-box manner. %

\noindent\textbf{Prompt Enhancement for Single-Label Recognition.} Several methods use LLM-generated descriptive prompts for single-label recognition, combining them via max or mean fusion \cite{CuPL,DCLIP,FuDD,CHiLS}. While related, our compound prompts specifically target MLR by modeling class relationships rather than enriching individual class descriptions.

\noindent \textbf{Complementarity of SPARC to Unsupervised Methods.} We explicitly note that our methodology is complementary to much of the work on unsupervised VLM-based MLR. For instance, since we assume only black-box access to VLM scores, our crude compound prompts can be refined via prompt tuning in the manner of TaI-DPT~\cite{TaI-DPT}, and can be applied at a patch-by-patch level and integrated \`{a} la TagCLIP~\cite{TagCLIP}. The resulting scores can then directly be combined via our debiasing and rank-fusion approach, and thus yield improvements to these prior approaches. \S\ref{sec:experiment_complementarity} investigates the resulting improvements.

\section{Method}
\label{sec:method}

We detail our method, SPARC (see Alg.~\ref{alg:sparc} and Fig.~\ref{fig:methodFigure}).

\begin{algorithm}
\caption{SPARC Pipeline}
\label{alg:sparc}
\begin{algorithmic}
\STATE \textbf{Input:} Images $\mathcal{I}$, Class Names $\mathcal{C}$
\STATE \textbf{Output:} Final scores $\zeta^t_i$ for each image $t$ and class $i$
\STATE $P \gets \text{GenerateCompoundPrompts}(\mathcal{C})$ \hfill \textit{(Details in Supp.)}
\FOR{$t \in \mathcal{I}$}
   \STATE $s^t_i \gets \text{GetVLMScores}(t, i)$ \hfill\textit{(Singleton scores)}
   \STATE $\forall p \in P,$ $s^t_p \gets \text{GetVLMScores}(t, p)$ \hfill\textit{(Compound scs.)}
   \STATE // \textit{(Debias scores (Eqs.~\ref{eqn:debias_images},\ref{eqn:debias_classes}))}
   \STATE $\tilde{s}^t_i \gets \text{ImageDebias}(s^t_i)$, $\tilde{s}^t_p \gets \text{ImageDebias}(s^t_p)$
   \STATE $\bar{s}^t_i \gets \text{PromptDebias}(\tilde{s}^t_i)$,
   $\bar{s}^t_p \gets \text{PromptDebias}(\tilde{s}^t_p)$
   \FOR{$i \in \mathcal{C}$}
       \STATE $r^t_{i,k} \gets \text{GetOrderStatistics}(\{\bar{s}^t_p: i \in p\})$
       \STATE $w^{i*} \gets \text{MaxVarDirection}([\bar{s}_i^t, r^t_{i,k}])$ \hfill 
 \textit{(Eq.~\ref{eq:maxVariance})}
       \STATE $\tilde{\zeta}^t_i \gets w^{i*}_0 \bar{s}^t_i + \sum_k w^{i*}_k r^t_{i,k}$ \hfill \textit{(Eq.~\ref{eqn:pca_projection})}
       \STATE $\zeta^t_i \gets \bar{s}^t_i + \tilde{\zeta}^t_i$  \hfill \textit{(Eq.~\ref{eq:merge})}
   \ENDFOR
\ENDFOR
\RETURN $\{\zeta^t_i\}$
\end{algorithmic}
\end{algorithm}

\noindent \emph{Setup.} %
Suppose we have $M$ images and $N$ target classes, with classnames $c_1,...,c_N$. We assume that we have a set of ``singleton'' prompts describing each of these classes in isolation (e.g. ``picture of a cat''). Given an image $t$, let $s_1^t,...,s_N^t$ denote the similarity scores produced by a VLM when comparing the singleton prompts to that image.

A na\"{i}ve approach is to directly use these singleton scores to perform MLR. However, since the presence of a particular class alters the conditional probability of the remaining classes in images, it should be possible to further refine these scores by accounting for the multi-label structure of the images. In order to do this, we use the classnames to generate a further set of ``compound'' prompts $P$ which mention multiple classes. For instance, the classes ``cat'' and ``sofa'' may be compounded to ``cat and sofa''. Let $\{s_p^t : p \in P\}$ denote the scores given by the VLM when comparing these compound prompts to image $t$.

Using the singleton as well as the compound prompts above, SPARC produces a vector of refined class-wise scores $\zeta_1^t, \cdots, \zeta_N^t,$ for each image $t$. Given this structure, SPARC has three main components: (i) compound prompt generation to enable the use of CLIP scoring for MLR (ii) image and prompt level debiasing to allow scores to be directly compared, and (iii) Rank Fusion to combine singleton and compound prompt scores into a single score per class. We now discuss each of these aspects in detail. 

\subsection{Compound Prompt Generation}
Our compound prompt generation method relies on contextual associations from commonly observed visual patterns to inform prompt structure. Specifically, we remove object pairs and triplets that are implausible based on low-frequency observations in the data, retaining only combinations that align with realistic visual scenes. Leveraging these contextual associations, an off-the-shelf LLM generates descriptive prompts that expand beyond the target class names, enriching the diversity of object scenarios. While we explored various prompt templates ("A OR B", "A next to B", "A with B"), "A and B" performed best in initial experiments. We focus on this template to establish our core method, leaving systematic prompting as future work.
Pseudocode and details are available in the Supplementary.

\subsection{Debiasing}
Once we have obtained compound prompts $P$, we can query the VLM to obtain singleton scores $s_1^t,...,s_N^t$ and compound scores $\{s_p^t : p \in P\}$ for each image $t$. However, these scores contain both image- and prompt-specific biases. The former in particular is a problem, as it can change the order of scores between negative and positive images.

We address image-level bias for singleton and compound prompts, respectively, as follows:
\begin{align}\label{eqn:debias_images}
\tilde{s}_i^t := \frac{s_i^t - \hat{\mu}(s_{\cdot}^t)}{\hat{\sigma}(s_{\cdot}^t)} \quad \textit{and} \quad
\tilde{s}_p^t := \frac{s_p^t - \hat{\mu}(\check{s}_{\cdot}^t)}{\hat{\sigma}(\check{s}_{\cdot}^t)},
\end{align}
where $\hat{\mu}(s_{\cdot}^t)$, $\hat{\mu}(\check{s}_{\cdot}^t)$, $\hat{\sigma}(s_{\cdot}^t)$, $\hat{\sigma}(\check{s}_{\cdot}^t)$ are sample means and standard deviations across the prompt dimension for a single image. $\check{s}_1^t,...,\check{s}_N^t$ are the scores of ``auxiliary'' prompts that only mention the classnames $c_1,...,c_N$ and are used only to obtain statistics for standardizing the compound prompt scores. We use these instead of singleton prompts for this purpose as the latter might have different statistical properties depending on the nature of the singleton prompting method.

We also standardize across images to quash prompt bias:
\begin{align}\label{eqn:debias_classes}
\overline{s}_i^t := \frac{\tilde{s}_i^t - \hat{\mu}(\tilde{s}_i^{\cdot})}{\hat{\sigma}(\tilde{s}_i^{\cdot})} \quad \textit{and} \quad
\overline{s}_p^t := \frac{\tilde{s}_p^t - \hat{\mu}(\tilde{s}_p^{\cdot})}{\hat{\sigma}(\tilde{s}_p^{\cdot})},
\end{align}
where $\hat{\mu}(\tilde{s}_i^{\cdot})$, $\hat{\mu}(\tilde{s}_p^{\cdot})$, $\hat{\sigma}(\tilde{s}_i^{\cdot})$, and $\hat{\sigma}(\tilde{s}_p^{\cdot})$ now denote sample means and SDs across the images. Removing prompt-level bias makes scores from different prompts more compatible with each other, which is important for fusing them.

\subsection{Rank Fusion}
Our goal now is to find a way to use the compound scores to strengthen the singleton scores. Let us define a bit more notation here. Let $C(p)$ be a function that maps each compound prompt to the set of classes in $[N]$ that are mentioned by that prompt. Let $r_{i,k}^t$ denote the $k$-th largest element of the set $\{\overline{s}_p^t : i \in C(p)\}$.

A natural choice might be to choose $r_{i,1}^t$, i.e. the highest scoring prompt. After all, we would expect this to be the prompt that most closely describes the image, and if the image did not contain $c_i$ then we would expect none of the prompts containing $c_i$ to score high. However we find that $r_{i,1}^t$ does not offer a good detection of classes. The mechanism behind this, which is explored in detail in \S\ref{sec:noisemodel}, is that when using a compound prompt of the form `A and B' to detect the class $A$, the score sees a large increase if the object $B$ truly occurs in the image. This `OR'-like behaviour (\S\ref{sec:noisemodel}) means that maximum score $r_{i,1}^t$ typically captures the effect of other classes than $A$ being present in the image, and thus leads to very poor separation between ground truth positive and negative images for any class. Counterintuitively, then, we find that `\emph{weakened maxes}', i.e., lower order statistics like $r_{i,2}^t$ and $r_{i,3}^t$ yield much better separation. 

The above fact necessitates that we need to deal with the entirety of the order statistics $\{r_{i,k}^t\}$ in order to generate an effective fusion rule. We fuse these scores by projecting the vector of order statistics along the direction of highest variance, i.e., we compute the fused compound score $\tilde{\zeta}_i^t$ as
\begin{align}
\label{eq:maxVariance}
w^{i*} &:= \argmax_{w^i} \textrm{Var}_t(w_0^i \bar{s}_i^t + \sum_{k} w_k^i r_{i,k}^t )\\
\tilde{\zeta}_i^t &:= w_0^{i*} \bar{s}_i^t + \sum_{k} w_k^{i*} r_{i,k}^t. \label{eqn:pca_projection}
\end{align}
Notice that the variance above is computed across all images. The main idea is similar to linear discriminant analysis. Indeed, if we assume that the mean compound score are when class $i$ is actually present or absent differ from one another, then the collection of the compound scores across all images can be modeled as a mixture of two noisy distributions, with some mean separation. If we further assume that the noise in these scores are roughly homoskedastic in either component, then, as long as the `signal' induced by the above mean shift is large compared to the noisy variation, highest variance direction $w^{i*}$ captures precisely the direction of the mean shift, and thus $\tilde{\zeta_i^t}$ effectively computes the component of the score vector that lies along this signal direction. While principled, this max-variance strategy is of course not necessary to execute, and we compare it to alternative fusion strategies in \S\ref{subsubsec:RankFusionAblation}. We add (i.e. ``merge'') this fused score into the original singleton score to get our final score:
\begin{equation}
\label{eq:merge}
\zeta_i^t := s_i^t + \tilde{\zeta}_i^t
\end{equation}
In the next section we probe CLIP scores through a statistical lens to build intuition on our approach.

\section{Noise Model of a VLM}
\label{sec:noisemodel}

Our methodological choices are strongly driven by the behaviour of CLIP scores under compound prompts, and this leads to both the debiasing approach taken, as well as our counterintuitive choice of exploiting the weakened maxes rather than just fusing the compound prompts via the maximum score. In order to justify these choices, we probe the structure of CLIP scores under compound prompts by constructing a simple model of the same.

Given an image $t$ and a prompt with classes $i$ and $j$, we model the score of compound prompt `$i$ and $j$' as \mbox{$s_{i,j}^t := \theta_1^t \cdot f(y_i^t, y_j^t) + \theta_0^t + \varepsilon$}
where $y_i^t, y_j^t \in \{0,1\}$ indicate ground-truth presence of of the two classes, $\theta_1^t$ and $\theta_0^t$ represent image-level bias (e.g. due to visual domain-shifts), and $\varepsilon$ is a zero-centered Gaussian noise.
Notice that our Debias module is well-suited to mitigate variations in $\theta_0^t$ and $\theta_1^t$, at least when $\varepsilon$ is not strongly dominating. 

Now, the core object of interest is the map $f : \{0,1\}^2 \to \mathbb{R},$ which is responsible for modeling class pair interactions and prompt-level biases (see Table~\ref{tab:NoiseModelTable}). In an ideal situation for MLR, $f$ would behave roughly as an `AND' function, so that compound prompts would let us directly hone in on the presence of pairs of classes. However, prior work has observed that instead, for CLIP-type models, $f$ tends to predominantly behave as an `OR' function, in that $f$ is high if either $i$ or $j$ is present. Notice that if $f$ were to exactly behave as an OR, compound prompts would not be effective in detecting classes that were missed by the singleton scores.

In experimentation, we found that the behaviour of $s_{i,j}^t$ is intermediary to the above extremes: CLIP scores for compound prompts behave qualitatively as an OR function with a small AND `bonus' when both classes are present. This bonus, as well as the particular scores, are modulated by the classes mentioned (a consequence of prompt-level bias for each class) which we model as linear effects, leading to an overall model of the signal term of the CLIP score as \begin{align}
f(y_i^t, y_j^t) &:= \max(a_0^i + y_i^t a_1^i, a_0^j + y_j^t a_1^j) \nonumber \\ &+ \delta_{ij} \min(a_0^i + y_i^t a_1^i, a_0^j + y_j^t a_1^j),
\end{align} where the $a$ terms are class-wise linear effect coefficients, and $\delta_{ij}$ captures the strength of this AND bonus (which depends on which pair is being queried).

Note that this AND bonus is a critical feature of CLIP scores in their use for MLR problems, since this bump allows us to both infer the presence of classes that singleton prompts have missed, and indirectly to filter false positives (since the true positives are further elevated by the bonus).

\begin{table}[t]
\centering
\begin{tabular}{| c | c | c |}
  \hline
  \textbf{Noise Model} & $f(y_i^t, y_j^t)$ & \textbf{FVU} \\ \hline
  constant  & $0.1$ & 0.802 \\ \hline
  only AND    & $\min(a_0^i\! +\! y_i^t a_1^i, a_0^j + y_j^t a_1^j)$    & 0.536 \\ \hline
  only OR    & $\max(a_0^i + y_i^t a_1^i, a_0^j + y_j^t a_1^j)$  & 0.288 \\ \hline
  additive   & $a_0^i + y_i^t a_1^i + a_0^j + y_j^t a_1^j$   & 0.267 \\ \hline
  \makecell{OR + static\\ AND-bonus}    & $\makecell{\max(a_0^i + y_i^t a_1^i, a_0^j + y_j^t a_1^j)+\\ \delta \min(a_0^i\! +\! y_i^t a_1^i, a_0^j\! +\! y_j^t a_1^j)}$  & 0.263 \\ \hline
  \makecell{OR + variable\\ AND-bonus}  & $\makecell{\max(a_0^i + y_i^t a_1^i, a_0^j + y_j^t a_1^j)+\\ \delta_{i,j} \min(a_0^i\! +\! y_i^t a_1^i, a_0^j\! +\! y_j^t a_1^j)}$  & 0.248 \\ \hline
  look-up table    & $\textrm{LUT}^{(i,j)}(y_i^t, y_j^t)$   & 0.235 \\ \hline
\end{tabular}
\vspace{-5pt}
\caption{Comparison of fidelity of noise models for scoring pairwise compound prompts. Notice that the OR+AND-bonus model captures nearly all of the fidelity of the look-up table, and offers significant improvements over the simplified models above it.}\vspace{-\baselineskip} %
\label{tab:NoiseModelTable}
\end{table}

Table~\ref{tab:NoiseModelTable} shows the results of a systematic evaluation of the AND-OR structure of the compound CLIP scores. We evaluate a variety of models of $f$, ranging from simple constant, `only OR,' and `only AND' models to additive models and OR+AND-bonus models with both a `static' $\delta$ shared by all class pairs, and a `variable' $\delta_{i,j}$ that is specific to each class pairs. The final entry is a `look-up table' (LUT) model wherein all four values of $f$ are allowed to vary arbitrarily, which serves as the most expressive baseline in this model class. These are evaluated w.r.t. the fraction of variance unexplained (i.e., $1-R^2,$ where $R$ is the residual). Observe that while the only OR model has remarkable fidelity, the additive and AND-bonus models significantly improve upon this, and allowing $\delta$ to vary with $i,j$ yield a further significant improvement to within a $6\%$ extra loss from the LUT model. This justifies our above model, and shows that it captures a significant amount of the qualitative behaviour of CLIP scores for compound prompts. 

A closer look at these fitted models reveals the $\delta$ values in these models are both small, and relatively stable: for the static $\delta,$ the recovered value is $0.56$, and the interquantile range of the varying $\delta_{ij}$ is $(0.44, 0.57)$.  %

We find similar behavior when examining other backbones and datasets, which we detail in the Supplementary.

We now use this noise model to explain the strengths of the weakened max from a theoretical perspective.

\noindent \textbf{Theoretical Explanation for Weakened Max.}
We observe empirically in Sec.~\ref{subsubsec:RankFusionAblation} that using the second-max score outperforms the first-max in discriminative ability. %
Consider a target class \(0\) with cooccurring classes \(1, \dots, m\) and assume that each cooccurrence is governed by a simple distribution with probabilities \(\rho\) and \(q\) as follows:
\[
    \Pr(y_i = 1 \mid y_0 = 1) = \rho, \quad \Pr(y_i = 1 \mid y_0 = 0) = q < \rho.
\]
Let us now assume that the scores for these compound prompts are derived from the following simplified noise model:
$s_{0,i} = \max(\tilde{y}_0, \tilde{y}_i) + \delta \min(\tilde{y}_0, \tilde{y}_i) + \varepsilon,$
where \(\varepsilon \sim \mathcal{W}(\sigma)\) is additive noise, and $\tilde{y}_0,\, \tilde{y}_i \in \{0,1\}$ are noisy, potentially flipped versions of ground truth $y_0, y_i,$  respectively. This label-flip noise models the effects of real-world conditions such as occlusion or partial visibility of each object in the input to the VLM. We informally state our theoretical finding. See Supplementary for full details:\\
\textbf{Theorem (informal)} For any distribution of label noise on \(y_0\), the second-max consistently improves discrimination over the first-max, given enough cooccurring classes.
\subsection{Methodological Implications}

The qualitative structure of the OR+AND-bonus behaviour observed in the previous section offers a robust justification for our methodological choices in \S\ref{sec:method}. Naturally, image level debiasing reduces the effect of variation in $\theta_0^t$ (via the mean subtraction) and $\theta_1^t$ (via the scaling by standard deviation). Further, prompt level debiasing reduces the effect of variations in $a_0$ and $a_1$ across classes. 

We can also qualitatively observe that due to the fact that the OR structure of the scores dominates the AND structure, the behaviour of the largest score $r_{i,1}^t$ should often reflect the presence of a related class $j$ instead of that of $i$, and suggest that subsequent order statistics are more informative. In fact, this property can be theoretically derived under mild assumptions, as stated above.

\section{Experiments}
\label{sec:experiments}

\paragraph{Datasets}
We benchmark our method on 3 different MLR datasets. They are COCO2014 \cite{COCO2014}, which has 40,504 test images with 80 classes; VOC2007 \cite{VOC2007}, which has 4.952 test images with 20 classes; and NUSWIDE \cite{nuswide}., which has 107,859 test images with 81 classes. We use only test, and not train, images from these datasets. We do use their conditional ground-truth label distributions, computed from the training set, to choose compound prompts, but note that our method requires only rough information about which classes are more or less likely to cooccur.\\

\noindent\textbf{CLIP Backbones}
In order to showcase the generality of our method, we apply it to 9 different CLIP backbones, which are ViT-L/14@336px, ViT-L/14, ViT-B/16, ViT-B/32, RN50x64, RN50x16, RN50x4, RN101, RN50. It is worth noting that from our model's perspective, each backbone could be seen as a dataset of its own, providing a different source of scores for each image set. We use the default 224x224 resolution, resizing without center crop, for all ViT models. For ResNet models, we use the same resolution as \cite{DualCoOp++} except when otherwise stated.

\subsection{Our method improves over vanilla ZSCLIP across many datasets and models}

\begin{table*}[ht]
\centering
\scalebox{0.95}{
\begin{tabular}{llccccccccc|c}
\toprule
\multirow{2}{*}{Dataset} & \multirow{2}{*}{} & \multicolumn{10}{c}{Architectures} \\
                         &                   & \makecell{ViT-L/14 \\ 336px} & ViT-L/14 & ViT-B/16 & ViT-B/32 & \makecell{RN50 \\ x64} & \makecell{RN50 \\ x16} & \makecell{RN50 \\ x4} & RN101 & RN50 & \makecell{Avg \\ (archs)} \\
\midrule
\multirow{2}{*}{COCO}    & ZSCLIP            & 59.1 & 58.1 & 55.5 & 50.9 & 58.6 & 58.1 & 55.7 & 52.7 & 53.0 & 55.7 \\
                         & Ours              & \textbf{70.5} & \textbf{69.5} & \textbf{67.4} & \textbf{64.0} & \textbf{70.6} & \textbf{70.1} & \textbf{69.5} & \textbf{67.7} & \textbf{65.7} & \textbf{68.3} \\
\midrule
\multirow{2}{*}{VOC}     & ZSCLIP            & 81.3 & 79.9 & 79.9 & 77.5 & 83.1 & 81.6 & 80.6 & 80.3 & 79.7 & 80.4 \\
                         & Ours              & \textbf{88.9} & \textbf{88.3} & \textbf{88.7} & \textbf{87.5} & \textbf{90.5} & \textbf{90.3} & \textbf{90.0} & \textbf{89.7} & \textbf{89.2} & \textbf{89.2} \\
\midrule
\multirow{2}{*}{NUSWIDE} & ZSCLIP            & 41.4 & 41.0 & 40.9 & 38.7 & 39.8 & 38.5 & 38.8 & 35.7 & 38.5 & 39.3 \\
                         & Ours              & \textbf{47.5} & \textbf{47.1} & \textbf{47.3} & \textbf{46.9} & \textbf{47.4} & \textbf{47.9} & \textbf{48.3} & \textbf{46.5} & \textbf{45.8} & \textbf{47.2} \\
\bottomrule
\end{tabular}
}
\vspace{-5pt}
\caption{\captionMainTableThreeDatasets}
\label{tab:MainTableThreeDatasets}
\end{table*}

\begin{figure}[t]
  \centering
    \includegraphics[width=1.05\linewidth]{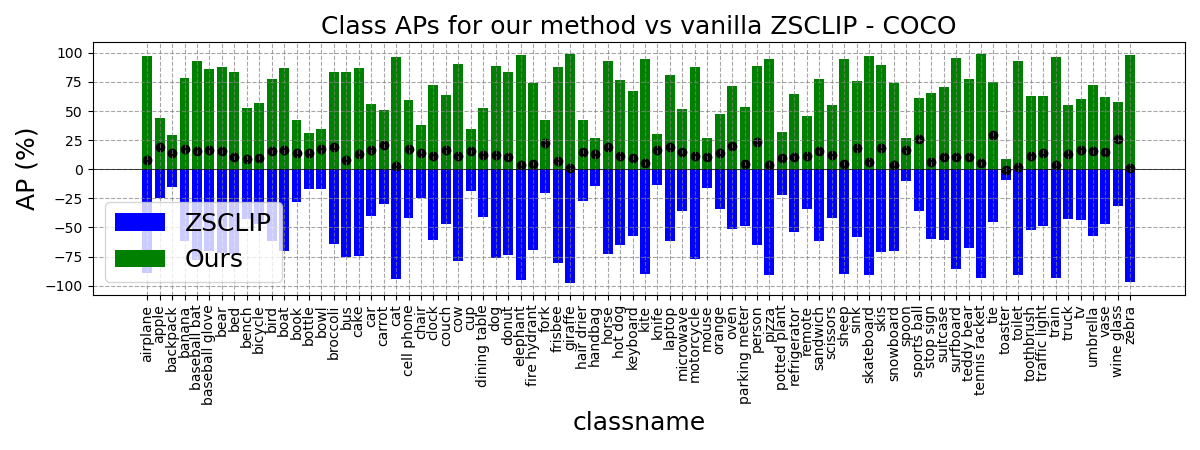}
    \vspace{-20pt}
   \caption{\captionPerClassAPsCOCO}
   \label{fig:PerClassAPsCOCO}
\end{figure}

As a baseline, we compute singleton scores using prompt-ensembling with the 80 templates from the CLIP paper \cite{CLIP}. We then compute compound prompt scores and apply our method to the singletons and compounds. We compare our results to this vanilla ZSCLIP baseline.

Tab.~\ref{tab:MainTableThreeDatasets} shows our results over three different datasets and nine different CLIP backbones. We see that our method is able improve over vanilla ZSCLIP by 12.6\% on COCO, 8.8\% on VOC, and 7.9\% on NUSWIDE. Moreover, we see that our method acheives consistent improvements across all datasets and backbones, showcasing its generality. For example, the improvements range from 11.4-15.0\% for COCO, 7.4-10\% for VOC, and 6.1-10.8\% for NUSWIDE. In fact, we see in Fig.~\ref{fig:PerClassAPsCOCO} that SPARC consistently improves the AP of almost every class in COCO. We show similar improvements on VOC and NUSWIDE in Supplementary.

\subsection{Our method has complementary strengths}\label{sec:experiment_complementarity}

\begin{table}[ht]
\vspace{-10pt}
\centering
\begin{tabular}{llcc|c}
\toprule
    &             &           COCO &            VOC &               \\
    &             &       ViT-B/16 &       ViT-B/16 &            Avg \\
\midrule
 & TagCLIP &           70.9 &           91.7 &           81.3 \\
    & $+$Ours &           \textbf{75.6} &           \textbf{92.3} &  \textbf{83.9} \\
\bottomrule
\end{tabular}
\vspace{-5pt}
\caption{\captionCompetitionTagCLIPlogYESrowcalibbaseNO}
\label{tab:CompetitionTagCLIPlogYESrowcalibbaseNO}
\vspace{-16pt}
\end{table}

\begin{table*}[ht]
\centering
\begin{tabular}{llcccccc|c||lcc|c}
\toprule
    &  & \multicolumn{2}{c}{COCO} & \multicolumn{2}{c}{VOC} & \multicolumn{2}{c}{NUSWIDE} &    & & \multicolumn{2}{c}{COCO} & \\
    & &  RN50 & RN50* & RN50 & RN50* & RN50 & RN50* & Avg & & RN50 & RN50* & Avg\\
\midrule
 & TaI-DPT & 65.1 & 68.2 & 88.5 & 88.0 & 46.2 & 43.4 & 66.6 & CoMC & 68.8 & \textbf{71.3} & \textbf{70.0} \\
    & $+$Ours &  \textbf{68.2} & \textbf{70.1} &  \textbf{90.3} &  \textbf{90.2} &  \textbf{46.6} & \textbf{44.6} & \textbf{68.3} & $+$Ours & \textbf{68.9} & 70.6 & 69.7 \\
\bottomrule
\end{tabular}
\vspace{-5pt}
\caption{\captionCompetitionTaIDPTandCoMC}
\label{tab:CompetitionTaIDPTandCoMC}
\end{table*}

Several existing methods adapt CLIP to MLR tasks in an image-free (\cite{TaI-DPT}, \cite{DataFreeMLR}, \cite{TaiPlusPlus}, \cite{CoMC}, \cite{TaI-Adapter}) or training-free (\cite{CLIPSurgery}, \cite{TagCLIP}) manner. These approaches lie outside our scope, as we focus on methods that are deployable `out-of-the-box' on new datasets and VLMs, without requiring dataset-specific training or modifications to the VLM. Our black-box method allows seamless 'plug-and-play' integration, using scores from existing methods in place of the vanilla ZSCLIP singleton scores.

We combined our method with three existing methods that have complete, publicly-available codebases: (1) TagCLIP \cite{TagCLIP}, a training-free method that benchmarks on COCO and VOC with a ViT-B/16 backbone; (2) TaI-DPT \cite{TaI-DPT}, an image-free method that requires training, which benchmarks on COCO, VOC, and NUSWIDE with a RN50 backbone; (3) CoMC \cite{CoMC}, also an image-free method that requires training, which has a publicly-available RN50-based model for COCO with a single training seed.

There are some nuances regarding image preprocessing and score postprocessing. TaI-DPT and CoMC use different image preprocessing settings than us - theirs are lower resolution and take a center crop. For a fair comparison, we use these same preprocessing settings when getting compound prompt scores to combine with TaI-DPT and CoMC. We also try running both our and their method with our preprocessing setting, which we denote as ``RN50*''. We do not change our settings for TagCLIP, as they rely on a less conventional preprocessing setting for ViT-B/16 that would be somewhat difficult to translate between methods. As for score postprocessing, TagCLIP uses softmax, which we have found to have a debiasing effect, so we omit our image-level debiasing on TagCLIP's scores, and take the log of their scores to make them compatible with ours. We make no modifications to postprocessing for TaI-DPT or CoMC.

We show the results of these plug-and-play combinations in Tab.~\ref{tab:CompetitionTagCLIPlogYESrowcalibbaseNO} and Tab.~\ref{tab:CompetitionTaIDPTandCoMC}. We find that method is able to improve on TagCLIP by on average 2.6\% and on TaI-DPT by on average 1.7\%. We do not improve on CoMC, but the degradation is only 0.3\%. These results suggest that our method's signal complements those obtained through training and architectural manipulation by other methods.

\subsection{Ablations}

\subsubsection{Debias Module}

\begin{table}[ht]
\centering
\scalebox{0.9}{
\begin{tabular}{cc|ccc|c}
\toprule
Compound &       Debias & COCO &  VOC & NUSWIDE &  Avg \\
\midrule
             &              & 55.7 & 80.4 &    39.3 & 58.5 \\
             & $\checkmark$ & 65.9 & 87.7 &    45.1 & 66.2 \\
$\checkmark$ &              & 58.4 & 80.5 &    40.0 & 59.7 \\
$\checkmark$ & $\checkmark$ & \textbf{68.3} & \textbf{89.2} & \textbf{47.2} & \textbf{68.3} \\
\bottomrule
\end{tabular}
}
\vspace{-5pt}
\caption{\captionDebiasAblationTable}
\label{tab:DebiasAblationTable}
\end{table}

We take a closer look at the Debias module by seeing how it impacts the performance of both singleton prompts and our full pipeline. We see in Tab.~\ref{tab:DebiasAblationTable} that the Debias module provides significant gains in both cases (7.7\% with singletons, 8.6\% full pipeline), confirming that image-level bias significantly impacts MLR when left unaddressed. %

\subsubsection{Rank Fusion Module}
\label{subsubsec:RankFusionAblation}
Our next ablation takes a closer look at the role of the Rank Fusion module. We consider some possible handcrafted strategies for combining compound prompt scores:
\begin{itemize}
    \item ``$k$max'': Use the $k$-th highest compound score, i.e. $r_{i,k}^t$
    \item ``mean$>=k$'': Use partial avg $\frac{1}{m_i-K+1} \sum_{k=K}^{m_i-K+1} r_{i,k}^t$
\end{itemize}
\begin{figure}[h]
  \centering
   \includegraphics[width=0.73\linewidth, trim=0 0 0 15, clip]{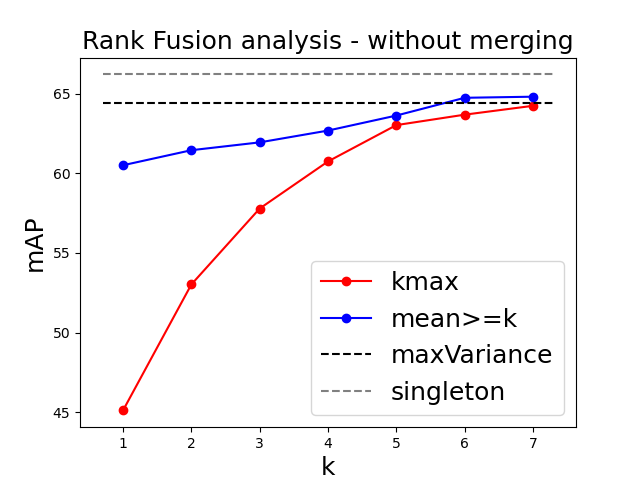}
   \includegraphics[width=0.73\linewidth, trim=0 0 0 15, clip]{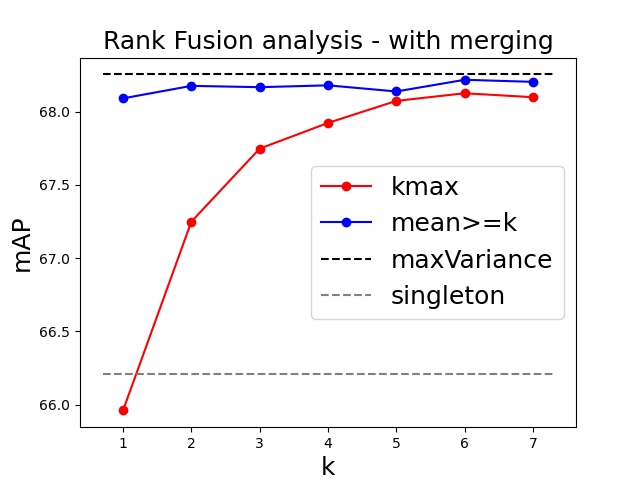}
   \caption{\captionRankFusionAblation}
   \label{fig:RankFusionAblation}
\end{figure}
Fig.~\ref{fig:RankFusionAblation} shows the average mAP of these strategies, as well as ``maxVariance'' \eqref{eq:maxVariance} and singleton-only, across all datasets and backbones, with and without the last ``merge'' step \eqref{eq:merge}. We see that the 1st-max does much worse than the 2nd-max, which does worse than the 3rd-max, and so on, up until a plateau somewhere around 5th-max. This lends credence to the idea that a weakened max is a more useful \mbox{signal} than outright max. We also see that the mean of all compound prompts (i.e. ``mean$>=\!1$'') does surprisingly well. This is actually consistent with our observation about weakened maxes - the mean weakens the 1st- and 2nd-max by averaging them with the lower ranks. We can improve further on the mean, to the point where we surpass the kmaxes, by excluding high ranks in the ``mean$>=k$'' strategy. However, we can surpass all of these options by adaptively weighting the scores using the max-variance direction.\\
Comparing the two figures also reveals that the ``merge'' step in \eqref{eq:merge} is critical and confirms that the score from ``maxVariance'' is indeed complementary to the singleton score.

We note that these results use CLIP's 80-prompt ensemble for singleton scores, indicating gains from compound prompting beyond basic prompt diversity.

\subsubsection{Compound Prompts}

The WaffleCLIP paper \cite{WaffleCLIP} found that prompt ensembles made of random texts could perform just as well as descriptive ensembles generated by LLMs, suggesting that the performance gains of the latter came not from semantics, but from the nature of ensembling itself. Given this result, we think it is always worth checking whether any prompt-enhancement method is actually benefiting from semantics.

To answer this question, we try using WaffleCLIP-style prompts in place of the compound prompts in our method. Specifically, for each classname $c_i$ we create 30 prompts of the form ``A photo of a $c_i$, which is [RAND]'', where ``[RAND]'' is 10 random characters. We try both the ``maxVariance'' strategy with merging and also ``mean$>=\!1$'' with merging, as the latter most closely resembles an ensemble.

We show the results of this ablation in the Supplementary. We see that randomized compound prompts do not offer any complementary signal to the singleton prompts. Thus, any gain from the compound prompts must come from their semantics. Just as our gains over random prompts come from semantics, they also extend beyond simple prompt diversity - our baseline singleton scores already incorporate CLIP's ensemble template.

\subsection{A qualitative look at the weakened max}
\label{subsec:qualitative}

\begin{figure}[t]
  \centering
   \includegraphics[width=0.8\linewidth]{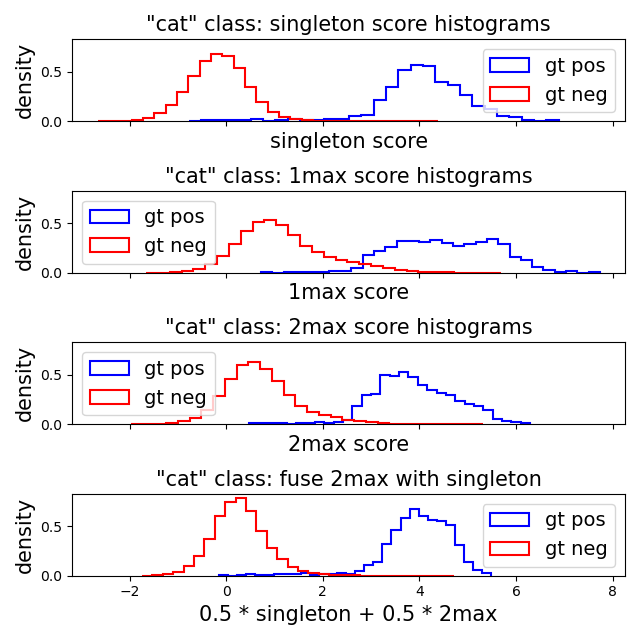}
   \caption{\captionQualitativeHist}
   \label{fig:QualitativeHist}
\end{figure}

We saw in Sec.~\ref{subsubsec:RankFusionAblation} that the second-highest compound prompt score is a much better signal than the highest one. This counterintuitive finding can be intuitively understood through object co-occurrence. When CLIP processes a compound prompt like 'cat and sofa', it has OR-like behavior - giving high scores if either object is present. The maximum score often comes from such prompts where only one object is present but strongly detected. In contrast, the second-highest scores tend to come from prompts where both objects have moderate confidence, providing a more reliable signal for true multi-label detection. This explains why fusing second-highest scores with singleton scores produces better separation between positive and negative cases.

Fig.~\ref{fig:QualitativeHist} shows the distributions of singleton, 1st-max, and 2nd-max scores for ``cat'' in COCO, as well as the uniform average of singleton and 2nd-max. We see in the second row that the 1st-max lifts a considerable number of negative examples, creating overlap between the negative and positive distributions. The third row shows that 2nd-max causes less overlap, lifting fewer negative examples without adversely impacting positive examples. The last row shows that fusing 2nd-max with singleton leads to good separation.

Fig.~\ref{fig:exampleFigure} shows an example of what happens at an image level. On the right is an image that contains a cat (curled up behind the two dogs on the bed); on the left is an image without a cat. The highest-scoring compound prompt is ``Cat and Dog'' for both images. This prompt gives a high score to the right image by detecting the dogs, but it also detects the dog in the left image for the same reason, and so it also misorders the images. But the second-highest-scoring prompts - ``Bed and Cat'' for right image, ``Cat and Potted plant'' for left - correctly order the images.

\section{Conclusions}
We presented SPARC, a zero-shot multi-label recognition approach that improves VLM performance without training data or architectural changes. Our key insights - the suboptimal nature of maximum scores and presence of systematic biases - led to two complementary innovations: compound prompts with adaptive fusion, and systematic debiasing. Beyond improving standalone VLM performance, SPARC enhances existing zero-shot and training-based methods while maintaining a black-box approach. The success of our method reveals fundamental properties of VLM scoring behavior, suggesting promising directions for improving zero-shot recognition through score analysis rather than architectural modification or prompt engineering.
\clearpage

{
    \small
    \bibliographystyle{ieeenat_fullname}
    \bibliography{main}
}

\clearpage
\setcounter{page}{1}
\maketitlesupplementaryonecolumn

\section{Supplementary Overview}
We organize our Supplementary Material as follows:
\begin{itemize}
    \item Sec.~\ref{sec:compound_prompt_generation_pseudocode} provides pseudocode for our compound prompt generation method.
    \item Sec.~\ref{sec:expanded_noise_model_results} extends the noise model analysis from Sec.~\ref{sec:noisemodel} of the main paper to all datasets and CLIP backbones.
    \item Sec.~\ref{sec:per_class_APs_all_datasets} shows the per-class performance of our method on all classes from all three datasets.
    \item Sec.~\ref{sec:compound_prompt_ablation_results} shows the results of different ablations on the compound prompts in our method, including randomized prompts, cooccurrence filtration, and compound prompt templates.
    \item Sec.~\ref{sec:theoretical_explanation_for_weakened_max} offers a theoretical justification for use of a ``weakened max'' and use of an adaptive fusion strategy. This includes a proof of the theorem that was informally stated at the end of Sec.~\ref{sec:noisemodel} in the main paper.
\end{itemize}

\section{Compound Prompt Generation Pseudocode}
\label{sec:compound_prompt_generation_pseudocode}

We provide pseudocode for our compound prompt generation method below. Note that it only requires \emph{coarse} knowledge of the cooccurrence probabilities, specifically knowledge of which pairs and triplets have low probability of cooccurring.

\begin{algorithm}[H]
\caption{Compound Prompt Generation}
\label{alg:compound_prompt_generation}
\begin{algorithmic}[1]
\REQUIRE Classnames $c_1, \ldots, c_N$, cooccurrence info $\mathbb{P}$, thresholds $\tau_2$ and $\tau_3$, optional LLM $\Phi$
\ENSURE Generated prompt set $P$

\STATE Initialize $P \leftarrow \emptyset$
\FOR{$i \in [N-1]$}
    \FOR{$j \in [i+1, N]$}
        \IF{$\mathbb{P}(j \mid i) > \tau_2$}
            \STATE $P \leftarrow P \cup \{\textrm{"$c_i$ and $c_j$"}\}$
            \IF{$\max_{k \in [N] - \{i,j\}} \mathbb{P}(k \mid i, j) > \tau_3$}
                \STATE $k^{*} \leftarrow \arg\max_{k \in [N] - \{i,j\}} \mathbb{P}(k \mid i, j)$
                \STATE $P \leftarrow P \cup \{\textrm{"$c_i$, $c_j$, and $c_{k^{*}}$"}\}$
            \ENDIF
        \ENDIF
    \ENDFOR
\ENDFOR
\STATE Remove redundant triplets $(i,j,k)$ vs $(i,k,j)$ by comparing $\mathbb{P}(k|i,j)$ and $\mathbb{P}(j|i,k)$
\IF{optional LLM $\Phi$ is provided}
    \STATE $P \leftarrow P \cup \Phi(P)$
\ENDIF
\RETURN $P$
\end{algorithmic}
\end{algorithm}

\section{Expanded Noise Model Results}
\label{sec:expanded_noise_model_results}

\begin{table*}[ht]
\centering
\begin{tabular}{l|lllllllll}
\toprule
{} & \multicolumn{9}{c}{COCO} \\
\textbf{Noise Model} & \makecell{ViT-L/14 \\ 336px} & ViT-L/14 & ViT-B/16 & ViT-B/32 & \makecell{RN50 \\ x64} & \makecell{RN50 \\ x16} & \makecell{RN50 \\ x4} &  RN101 &   RN50 \\
\midrule
constant                   &        0.802 &   0.812 &   0.823 &   0.775 &   0.809 &   0.798 &  0.757 &  0.771 &  0.791 \\
\hline
only AND                     &        0.535 &   0.544 &   0.557 &   0.517 &   0.553 &   0.539 &  0.501 &  0.506 &  0.500 \\
\hline
only OR                      &        0.288 &   0.301 &   0.292 &   0.295 &   0.284 &   0.281 &  0.257 &  0.260 &  0.280 \\
\hline
additive                     &        0.267 &   0.279 &   0.276 &   0.278 &   0.269 &   0.264 &  0.244 &  0.252 &  0.263 \\
\hline
\makecell{OR + static \\ AND-bonus}            &        0.263 &   0.275 &   0.271 &   0.273 &   0.264 &   0.260 &  0.239 &  0.245 &  0.257 \\
\hline
\makecell{OR + variable \\ AND-bonus}  &        0.248 &   0.257 &   0.258 &   0.259 &   0.245 &   0.247 &  0.228 &  0.233 &  0.245 \\
\hline
look-up table                 &        0.235 &   0.243 &   0.244 &   0.245 &   0.230 &   0.233 &  0.217 &  0.221 &  0.231 \\
\hline \hline
static bonus strength                        &        0.560 &   0.558 &   0.522 &   0.502 &   0.560 &   0.549 &  0.506 &  0.445 &  0.484 \\
\hline
\makecell{variable bonus strength \\ (lower quartile)}   &        0.422 &   0.408 &   0.386 &   0.366 &   0.433 &   0.405 &  0.389 &  0.348 &  0.381 \\
\hline
\makecell{variable bonus strength \\ (upper quartile)}    &        0.531 &   0.527 &   0.503 &   0.489 &   0.582 &   0.666 &  0.656 &  0.631 &  0.583 \\
\bottomrule
\end{tabular}
\caption{Comparison of fidelity of noise models for scoring pairwise compound prompts on COCO for all CLIP backbones. Notice that the OR-only model is a significantly better fit than AND-only, and that the OR+AND-bonus models capture nearly all of the fidelity of the look-up table. Please refer to Tab.~\ref{tab:NoiseModelTable} for more details on the noise models.}
\label{tab:NoiseModelExtendedCOCO}
\end{table*}

\begin{table*}[ht]
\centering
\begin{tabular}{l|lllllllll}
\toprule
{} & \multicolumn{9}{c}{VOC} \\
\textbf{Noise Model} & \makecell{ViT-L/14 \\ 336px} & ViT-L/14 & ViT-B/16 & ViT-B/32 & \makecell{RN50 \\ x64} & \makecell{RN50 \\ x16} & \makecell{RN50 \\ x4} &  RN101 &   RN50 \\
\midrule
constant                  &        0.640 &   0.648 &   0.661 &   0.616 &   0.603 &   0.631 &  0.619 &  0.628 &  0.653 \\
\hline
only AND                     &        0.275 &   0.276 &   0.275 &   0.282 &   0.276 &   0.299 &  0.245 &  0.230 &  0.242 \\
\hline
only OR                      &        0.144 &   0.150 &   0.143 &   0.136 &   0.137 &   0.127 &  0.123 &  0.122 &  0.126 \\
\hline
additive                     &        0.128 &   0.133 &   0.127 &   0.123 &   0.119 &   0.119 &  0.111 &  0.112 &  0.114 \\
\hline
\makecell{OR + static \\ AND-bonus}            &        0.125 &   0.130 &   0.125 &   0.120 &   0.118 &   0.116 &  0.110 &  0.111 &  0.111 \\
\hline
\makecell{OR + variable \\ AND-bonus} &        0.120 &   0.123 &   0.112 &   0.113 &   0.109 &   0.111 &  0.105 &  0.104 &  0.107 \\
\hline
look-up table                 &        0.114 &   0.116 &   0.109 &   0.109 &   0.104 &   0.105 &  0.100 &  0.099 &  0.101 \\
\hline\hline
static bonus strength                        &        0.427 &   0.412 &   0.459 &   0.408 &   0.557 &   0.367 &  0.356 &  0.441 &  0.493 \\
\hline
\makecell{variable bonus strength \\ (lower quartile)}    &        0.306 &   0.274 &   0.296 &   0.254 &   0.429 &   0.257 &  0.197 &  0.330 &  0.361 \\
\hline
\makecell{variable bonus strength \\ (upper quartile)}    &        0.582 &   0.555 &   0.873 &   0.419 &   0.680 &   0.461 &  0.321 &  0.833 &  0.792 \\
\bottomrule
\end{tabular}
\caption{Comparison of fidelity of noise models for scoring pairwise compound prompts on VOC for all CLIP backbones. Notice that the OR-only model is a significantly better fit than AND-only, and that the OR+AND-bonus models capture nearly all of the fidelity of the look-up table. Please refer to Tab.~\ref{tab:NoiseModelTable} for more details on the noise models.}
\label{tab:NoiseModelExtendedVOC}
\end{table*}

\begin{table*}[ht]
\centering
\begin{tabular}{l|lllllllll}
\toprule
{} & \multicolumn{9}{c}{NUSWIDE} \\
\textbf{Noise Model} & \makecell{ViT-L/14 \\ 336px} & ViT-L/14 & ViT-B/16 & ViT-B/32 & \makecell{RN50 \\ x64} & \makecell{RN50 \\ x16} & \makecell{RN50 \\ x4} &  RN101 &   RN50 \\
\midrule
constant                   &        0.536 &   0.531 &   0.562 &   0.581 &   0.565 &   0.557 &  0.534 &  0.534 &  0.589 \\
\hline
only AND                     &        0.330 &   0.331 &   0.346 &   0.358 &   0.351 &   0.352 &  0.344 &  0.350 &  0.371 \\
\hline
only OR                      &        0.228 &   0.230 &   0.234 &   0.248 &   0.251 &   0.239 &  0.222 &  0.223 &  0.256 \\
\hline
additive                     &        0.196 &   0.199 &   0.200 &   0.211 &   0.213 &   0.207 &  0.195 &  0.201 &  0.226 \\
\hline
\makecell{OR + static \\ AND-bonus}            &        0.193 &   0.195 &   0.196 &   0.207 &   0.210 &   0.204 &  0.191 &  0.197 &  0.222 \\
\hline
\makecell{OR + variable \\ AND-bonus} &        0.175 &   0.177 &   0.183 &   0.190 &   0.187 &   0.185 &  0.180 &  0.186 &  0.208 \\
\hline
look-up table                 &        0.162 &   0.164 &   0.170 &   0.177 &   0.173 &   0.171 &  0.167 &  0.173 &  0.194 \\
\hline\hline
static bonus strength                        &        0.604 &   0.603 &   0.579 &   0.584 &   0.637 &   0.597 &  0.581 &  0.556 &  0.591 \\
\hline
\makecell{variable bonus strength \\ (lower quartile)}    &        0.488 &   0.485 &   0.471 &   0.457 &   0.482 &   0.456 &  0.435 &  0.394 &  0.418 \\
\hline
\makecell{variable bonus strength \\ (upper quartile)}    &        0.647 &   0.619 &   0.626 &   0.620 &   0.644 &   0.624 &  0.663 &  0.628 &  0.609 \\
\bottomrule
\end{tabular}
\caption{Comparison of fidelity of noise models for scoring pairwise compound prompts on NUSWIDE for all CLIP backbones. Notice that the OR-only model is a significantly better fit than AND-only, and that the OR+AND-bonus models capture nearly all of the fidelity of the look-up table. Please refer to Tab.~\ref{tab:NoiseModelTable} for more details on the noise models.}
\label{tab:NoiseModelExtendedNUSWIDE}
\end{table*}

We use this section to show the results of running the noise model analysis from Tab.~\ref{tab:NoiseModelTable} on CLIP similarity scores from all three datasets computed with all nine CLIP backbones. Each colummn of each of Tab.~\ref{tab:NoiseModelExtendedCOCO}, Tab.~\ref{tab:NoiseModelExtendedVOC}, Tab.~\ref{tab:NoiseModelExtendedNUSWIDE} represents a separate fit of the noise models. As in the main paper, we report fraction of variance unexplained (FVU), as well as the fitted $\delta$ strength of the static AND-bonus and the lower and upper quartiles of the $\delta_{i,j}$ strengths of the variable AND-bonus. We see similar trends to those discussed in the main paper; the OR-only noise model explains significantly more variance than the AND-only model, and the OR-with-AND-bonus models explain most of the variance gap between the ``constant'' upper-bound and ``look-up table'' lower-bound. Hence, we find that CLIP scores tend to behave like an OR-gate with an AND-gate ``bonus'' for many different backbones on multiple datasets.

\section{Per-class performance of SPARC vs vanilla ZSCLIP on all datasets}
\label{sec:per_class_APs_all_datasets}

\begin{figure}[t]
  \centering
    \includegraphics[width=1\linewidth]{sec/figs/COCO2014_test_class_APs.png}
    \includegraphics[width=1\linewidth]{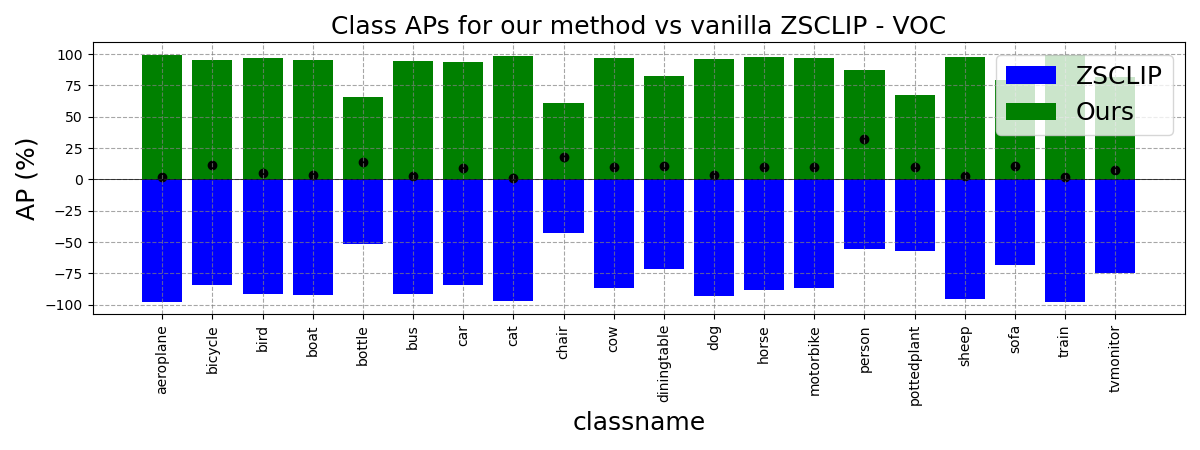}
    \includegraphics[width=1\linewidth]{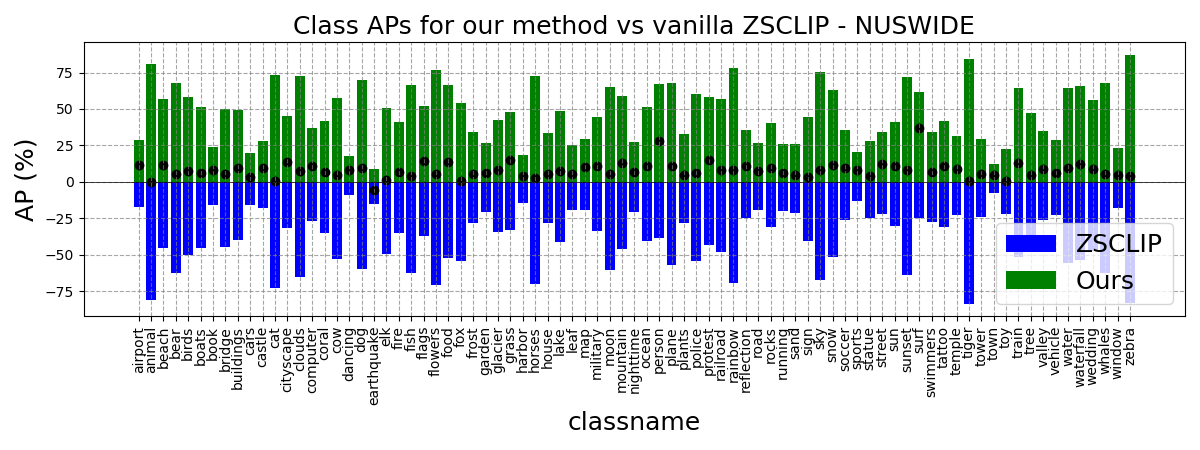}
    \caption{Per-class APs (averaged over all CLIP backbones) for our method vs vanilla ZSCLIP on all three datasets. Our method consistently improves over ZSCLIP for almost every class in all datasets.}
    \label{fig:PerClassAPsAllDatasets}
\end{figure}

We showed in Fig.~\ref{fig:PerClassAPsCOCO} in the main paper that SPARC \emph{consistently} improves over vanilla ZSCLIP across all the classes in COCO. We now show in Fig.~\ref{fig:PerClassAPsAllDatasets} that the improvement is consistent across all the classes in all three datasets. In fact, we see that there is only one class out of all the classes in all the datasets (181 classes total) where our method does notably worse than ZSCLIP. That class is the ``earthquake'' class of the NUSWIDE dataset. For all other classes, our method is either the same or (in most cases) notably better than ZSCLIP. 

\section{Compound prompt ablation results}
\label{sec:compound_prompt_ablation_results}

\begin{table*}[t]
\centering
\begin{tabular}{cccc|ccc|c}
\toprule
Use compound &       Debias & Compound prompt type & Rank Fusion Strategy & COCO &  VOC & NUSWIDE &  Avg \\
\midrule
             & $\checkmark$ &                    - &                    - & 65.9 & 87.7 &    45.1 & 66.2 \\
$\checkmark$ & $\checkmark$ &               randomized &                 ours & 65.9 & 87.5 &    45.1 & 66.2 \\
$\checkmark$ & $\checkmark$ &               randomized &                 mean & 65.9 & 87.4 &    45.1 & 66.1 \\
$\checkmark$ & $\checkmark$ &                 ours &                 ours & \textbf{68.3} & \textbf{89.2} &    \textbf{47.2} & \textbf{68.3} \\
\bottomrule
\end{tabular}
\caption{Randomized compound prompt ablation confirms the semantic value of compound prompts. Our ablation replaces compound prompts with prompts that use random characters instead of cooccurrent classes. These prompts offer no benefit over debiased singletons, suggesting that the gain caused by cooccurrent classes is due to semantics, and not just the statistical properties of an ensemble.}
\label{tab:WaffleCLIPAblationTable}
\end{table*}

We describe a few different ablations on the compound prompts used by our method. We start by comparing the performance of our compound prompts with ``randomized'' prompts in which the cooccurrent classes are replaced by random characters. We do this in light of the findings of WaffleCLIP \cite{WaffleCLIP}, which found that randomized prompt ensembles could perform as well as descriptive ones due to the inherent statistical benefits of ensembling. We find that this is not the case for our problem - randomized compound prompts offer no benefit over debiased singletons. We show our results in Tab.~\ref{tab:WaffleCLIPAblationTable}.

\begin{table*}[ht]
\centering
\begin{tabular}{cccc|ccc|c}
\toprule
Debias & Pair prompts & Triplets + Descriptive & Rank Fusion Strategy & COCO & VOC & NUSWIDE & Avg  \\
\midrule
$\checkmark$ & - & & - & 65.9 & 87.7 & 45.1 & 66.2 \\
$\checkmark$ & all pairs & & ours & 67.5 & 88.5 & 46.4 & 67.5 \\
$\checkmark$ & all pairs & & mean & 67.5 & 88.5 & 46.4 & 67.5 \\
$\checkmark$ & cooccurrence-filtered & & ours & 68.1 & 89.0 & 47.0 & 68.0 \\
$\checkmark$ & cooccurrence-filtered & & mean & 67.9 & 88.5 & 46.8 & 67.7 \\
$\checkmark$ & cooccurrence-filtered & $\checkmark$ & ours & $\textbf{68.3}$ & $\textbf{89.2}$ & $\textbf{47.2}$ & $\textbf{68.3}$ \\
\bottomrule
\end{tabular}
\caption{Ablations on the makeup of our compound prompts. We find that we can still enjoy some benefit from pairwise prompts without any cooccurrence filtering.}
\end{table*}

Our next ablation takes a closer look at the \textit{composition} of our set of compound prompts. We start by including \textit{all} pairs of classes in our set of formulaic pairwise prompts (i.e. of the form ``A and B'') and not including any other kind of compound prompt. We find that this gives us a 1.3\% average boost over debiased singletons. Filtering these pairwise prompts by cooccurrence, as our proposed method does, yields a further 0.5\% boost, and adding triplet and descriptive prompts gives an additional 0.3\% boost. It makes sense that using the full set of class pairs would still have some benefit, without any cooccurrence filtering, if all the classes as a whole tend to cooccur positively more often than negatively, as that would mean that they are on average a helpful signal for predicting each other's presence. It might also be the case that related classes provide a useful context signal to CLIP.

\begin{table*}[ht]
\centering
\begin{tabular}{ccc|ccc|c}
\toprule
Debias & Pair prompt template & Triplets + Descriptive & COCO & VOC & NUSWIDE & Avg  \\
\midrule
$\checkmark$ & - & & 65.9 & 87.7 & 45.1 & 66.2 \\
$\checkmark$ & ``A and B'' & & 68.1 & 89.0 & 47.0 & 68.0 \\
$\checkmark$ & ``A or B'' & & 67.0 & 88.4 & 46.2 & 67.2 \\
$\checkmark$ & ``A with B'' & & 67.8 & 89.0 & 47.0 & 67.9 \\
$\checkmark$ & ``A next to B'' & & 67.9 & 88.8 & 46.7 & 67.8 \\
$\checkmark$ & ``A and not B'' & & 67.9 & 88.7 & 46.5 & 67.7 \\
$\checkmark$ & all templates & & 67.9 & 88.7 & 46.7 & 67.8 \\
$\checkmark$ & ``A and B'' & $\checkmark$ & $\textbf{68.3}$ & $\textbf{89.2}$ & $\textbf{47.2}$ & $\textbf{68.3}$ \\
\bottomrule
\end{tabular}
\caption{Ablations on templates used for formulaic pairwise prompts. We find that our original template ``A and B'' performs best.}
\label{tab:compoundPromptTemplateAblation}
\end{table*}

Finally, we consider alternative templates for the formulaic pair prompts. In addition to the ``A and B'' template used in the main paper, we also try ``A or B'', ``A with B'', ``A next to B'', ``A and not B'' (alongside ``A and B''), and the combination of all templates. For simplicity, we remove the triplet and descriptive compound prompts during this analysis. We report the results in Tab.~\ref{tab:compoundPromptTemplateAblation}. We find that our original template ``A and B'' performs the best, although other conjunctive templates do get quite close, while ``A or B'' does considerably worse. This latter finding suggests that perhaps CLIP does interpret ``and'' and ``or'' differently, even if it treats ``A and B'' primarily as an OR-gate. 

\section{Theoretical Explanation for Weakened Max and Adaptive Fusion}
\label{sec:theoretical_explanation_for_weakened_max}

\subsection{Theory Overview}

We introduced a theoretical justification at the end of Sec.~\ref{sec:noisemodel} for the use of a ``weakened max'' instead of an outright maximum of compound scores. We will now explain that justification in full detail.

Recall that the goal was to predict the presence or absence of target class $0$ given a set of $m$ compound prompts pairing class $0$ with each of cooccurring classes $1,...,m$. We informally stated in Sec.~\ref{sec:noisemodel} that the second-max will outperform the first-max for sufficiently large $m$. We define our settings and assumptions more precisely in Sec.~\ref{subsec:theory_preliminaries} and then formally state this theorem as Theorem \ref{thm:theorem1}, which we prove in Sec.~\ref{sec:theorem1_proof}.

We make a further claim in Theorem \ref{thm:theorem2}, which states that there are settings for which a sufficiently \emph{small} $m$ will cause the first-max to outperform the second-max, and that the boundary between ``sufficiently large'' and ``sufficiently small'' depends on data statistics that are unknowable in any practical setting, even one where exact cooccurrence statistics are available. We prove this theorem in Sec.~\ref{sec:theorem2_proof}. We suspect that we would find similar behavior for other pairs of statistics, such as second-max vs third-max, third vs fourth, fourth vs median, median vs min, etc. In general, \textbf{fixed fusion rules are suboptimal} for combining the compound prompt scores.

Together, these theorems justify not only the use of a \textbf{``weakened max''}, but also the use of an \textbf{adaptive fusion strategy} such as Rank Fusion, which can use the direction of highest variance to figure out which order statistics are most useful for the setting at hand.

\subsection{Preliminaries}
\label{subsec:theory_preliminaries}

Suppose we have target class $0$ and cooccurring classes $1,..,m$. These have ground-truth presences $y_0, y_1,...,y_m \in \{0,1\}$. We make some assumptions about their distribution.

\begin{assumption}
\label{assumption:assumption1}
The ground-truth distribution has the following properties:
\begin{align}
Pr(y_i = 1 \ |\  y_0 = 1) &= \rho \qquad\quad \forall i \in [m]\\
Pr(y_i = 1\ |\ y_0 = 0) &= q \qquad\quad \forall i \in [m]\\
1 > \rho &> q > 0\\
y_j &\perp y_i\ |\ y_0\qquad\!\!\!\!\!\!\!\ \forall i \neq j \in [m]
\end{align}
\end{assumption}

We also introduce variables $\tilde{y}_0, \tilde{y}_1,...,\tilde{y}_m \in \{0,1\}$ which are ``noisy'' versions of the ground-truth. Think of these as indicating whether each object is visible to the VLM. E.g. we might have $y_i = 1$ and $\tilde{y}_i = 0$ if object $i$ was occluded, or we might have $y_i = 0$ and $\tilde{y}_i = 1$ if an spurious object in the image resembled $i$. For each $i \in [m]$ we have:
\[
\tilde{y}_i = 
\begin{cases} 
1 - y_i & \text{with probability } \nu, \\
y_i & \text{with probability } 1 - \nu.
\end{cases}
\]

We make some assumptions about the distribution of these variables.

\begin{assumption}
\label{assumption:assumption2}
The distribution of $\tilde{y}_0, \tilde{y}_1,...,\tilde{y}_m$ has the following properties:
\begin{align}
\tilde{y}_i & \ \textrm{depends only on}\ y_i\\
\nu &< \frac{1}{2}
\end{align}
\end{assumption}

As mentioned on the main paper, we assume that the score for the prompt ``$\{0\}$ and $\{i\}$'' is distributed as follows:
\begin{align}
s_{0,i} &= \max(\tilde{y}_0, \tilde{y}_i) + \delta \min(\tilde{y}_0, \tilde{y}_i) + \varepsilon
\end{align}
where $\delta$ is the strength of the ``AND-bonus'' described in the main paper, and $\varepsilon \sim \mathcal{W}(\sigma)$ is symmetric, zero-centered, additive noise whose scale is controlled by $\sigma$.\\

From the set $\{s_{0,1},...,s_{0,m}\}$ we compute order statistics $r_1, r_2$, which are the first and second highest elements, respectively.\\

Now, suppose we independently draw a ground-truth positive sample with $y_0^{+} = 1$ and a ground-truth negative sample with $y_0^{-} = 0$ and compute order statistics $r_1^{+}, r_2^{+}$ and $r_1^{-}, r_2^{-}$. We define "win" events $W_1$ and $W_2$ as the events where $r_1^{+} > r_1^{-}$ and $r_2^{+} > r_2^{-}$, respectively.\\

We will now make an assumption about $\varepsilon \sim \mathcal{W}(\sigma)$ in order to simplify our analysis. In order to state our assumption, we will need a bit more notation.
\begin{align}
\bar{s}_{0,i} &= \max(\tilde{y}_0, \tilde{y}_i) + \delta \min(\tilde{y}_0, \tilde{y}_i)\\
\bar{r}_1, \bar{r}_2\ &\textrm{are the first and second highest elements of}\ \{\bar{s}_{0,1},...,\bar{s}_{0,m}\}
\end{align}

We are now ready to state our assumption.
\begin{assumption}
\label{assumption:assumption3}
Assume that $\sigma$ is small enough for the following to approximately hold for each $k \in \{1,2\}$
\[
\textrm{Pr}(W_k) \approx 
\begin{cases} 
1 & \text{if }\ \bar{r}_k^{+} > \bar{r}_k^{-}, \\
0 & \text{if }\ \bar{r}_k^{+} < \bar{r}_k^{-}, \\
\frac{1}{2} & \text{if }\ \bar{r}_k^{+} = \bar{r}_k^{-},
\end{cases}
\]
\end{assumption}

We have now stated all of our assumptions.\\

Before making our formal theorem statements, we define some shorthand that we will use throughout the proof. First, we note that if we hold $\tilde{y}_0$ fixed, then $\bar{r}_1$ and $\bar{r}_2$ can each take on one of two values. For example, if $\tilde{y}_0 = 0$ then the possible values are $\{0,1\}$, and if $\tilde{y}_0 = 1$ then the possible values are $\{1, 1 + \delta\}$. As such, we define pairs of complementary events $(H_1, L_1)$ and $(H_2, L_2)$ to denote that $\bar{r}_1$ or $\bar{r}_2$ took the higher or lower of its possible values.\\

We define some additional shorthand:
\begin{align}
\rho^{\prime} &:= (1 - \nu) \rho\ +\ \nu (1 - \rho)\\
q^{\prime} &:= (1 - \nu) q\ +\ \nu (1 - q)\\
a &:= 1 - \rho^{\prime}\\
\gamma &:= \frac{1 - q}{1 - \rho^{\prime}}\\
A &:= m (1 - a) a^{m-1}\\
G &:= m (1 - \gamma a) (\gamma a)^{m-1}
\end{align}

We are now ready to formally state our theorems.

\subsection{Formal Theorem Statements}

\begin{theorem}
\label{thm:theorem1}
Given the assumptions above, plus the additional assumption that $\textrm{Pr}(\tilde{y}_0^{+} \neq y_0^{+} \bigvee \tilde{y}_0^{-} \neq y_0^{-}) > 0$, we can guarantee that $\textrm{Pr}(W_2) > \textrm{Pr}(W_1)$ for sufficiently large $m$.
\end{theorem}

\begin{theorem}
\label{thm:theorem2}
There are values of $\rho, q, \nu$ and distributions of $(\tilde{y}_0^{+}, \tilde{y}_0^{-})$ which satisfy all the requirements of Theorem \ref{thm:theorem1}, for which $\textrm{Pr}(W_2) < \textrm{Pr}(W_1)$ for sufficiently small $m$. In fact, the value of $m$ at which the inequality reverses depends on label-flip probability $\nu$.
\end{theorem}

\subsection{Proof of Theorem 1}
\label{sec:theorem1_proof}

We start by proving that $\rho^{\prime} > q^{\prime}$, i.e. $\textrm{Pr}(\tilde{y}_i = 1\ |\ y_0 = 1) > \textrm{Pr}(\tilde{y}_i = 1\ |\ y_0 = 0)$.

\begin{lemma}
\label{lemma:lemma1}
$1 > \rho^{\prime} > q^{\prime} > 0$ given the above assumptions on $\rho$, $q$, and $\nu$.
\end{lemma}

\begin{proof}
We can use some algebra to prove this from Assumptions~\ref{assumption:assumption1} and~\ref{assumption:assumption2}.
\begin{align}
\rho^{\prime} &= \rho + \nu - 2 \nu \rho\\
q^{\prime} &= q + \nu - 2 \nu q\\
\rho^{\prime} - q^{\prime} &= 2 (\frac{1}{2} - \nu) (\rho - q)\\
&> 0
\end{align}
It is trivial to show that $\rho^{\prime}, q^{\prime} \in (0,1)$ because they are both mixtures of quantities in that range.
\end{proof}

Our next lemma will derive some probability differences that will be important for our proof.

\begin{lemma}
\label{lemma:lemma2}
Consider the following probability differences:
\begin{align}
d^{HH} &:= \textrm{Pr}(H_2^{+}, H_2^{-}) - \textrm{Pr}(H_1^{+}, H_1^{-})\\
d^{HL} &:= \textrm{Pr}(H_2^{+}, L_2^{-}) - \textrm{Pr}(H_1^{+}, L_1^{-})\\
d^{LH} &:= \textrm{Pr}(L_2^{+}, H_2^{-}) - \textrm{Pr}(L_1^{+}, H_1^{-})\\
d^{LL} &:= \textrm{Pr}(L_2^{+}, L_2^{-}) - \textrm{Pr}(L_1^{+}, L_1^{-})
\end{align}
We claim that, regardless of the values or distribution of $\tilde{y}_0^{+}$ and $\tilde{y}_0^{-}$, the following is true:
\begin{align}
d^{HH} &= AG - (1 - a^m) G - (1 - (\gamma a)^m) A\\
d^{HL} &= (1 - a^m) G - AG - (\gamma a)^m A\\
d^{LH} &=  (1 - (\gamma a)^m) A - AG - a^m G\\
d^{LL} &= AG + a^m G + (\gamma a)^m A
\end{align}
\end{lemma}

\begin{proof}
We start by noting that $d^{HH}, d^{HL}, d^{LH}, d^{LL}$ do not depend on $\tilde{y}_0^{+}, \tilde{y}_0^{-}$. Although $\tilde{y}_0^{+}$ and $\tilde{y}_0^{-}$ affect the specific values that $\bar{r}_1^{+}, \bar{r}_2^{+}, \bar{r}_1^{-}, \bar{r}_2^{-}$ can take, they do not affect the probabilities of events $H_1^{+}, L_1^{+}, H_2^{+}, L_2^{+}, H_1^{-}, L_1^{-}, H_2^{-}, L_2^{-}$. This is because these events only depend on $\tilde{y}_1^{+},...,\tilde{y}_m^{+}$ and $\tilde{y}_1^{-},...,\tilde{y}_m^{-}$, which in turn depend on ground truths $y_1^{+},...,y_m^{+}$ and $y_1^{-},...,y_m^{-}$, which all depend on $y_0^{+}$ and $y_0^{-}$, which are fixed, so there is no dependency on $\tilde{y}_0^{+}$ or $\tilde{y}_0^{-}$.

We also note that we can factor out the joint probabilities in $d^{HH}, d^{HL}, d^{LH}, d^{LL}$ because the positive and negative samples were drawn independently, hence:
\begin{align}
d^{HH} &= \textrm{Pr}(H_2^{+}) \textrm{Pr}(H_2^{-}) - \textrm{Pr}(H_1^{+}) \textrm{Pr}(H_1^{-})\\
d^{HL} &= \textrm{Pr}(H_2^{+}) \textrm{Pr}(L_2^{-}) - \textrm{Pr}(H_1^{+}) \textrm{Pr}(L_1^{-})\\
d^{LH} &= \textrm{Pr}(L_2^{+}) \textrm{Pr}(H_2^{-}) - \textrm{Pr}(L_1^{+}) \textrm{Pr}(H_1^{-})\\
d^{LL} &= \textrm{Pr}(L_2^{+}) \textrm{Pr}(L_2^{-}) - \textrm{Pr}(L_1^{+}) \textrm{Pr}(L_1^{-})
\end{align}

We can work out the probability for event $H_1^{+}$, which occurs iff at least one of $\tilde{y}_1^{+},...,\tilde{y}_m^{+}$ is $1$:
\begin{align}
\textrm{Pr}(H_1^{+}) &= 1 - (1 - \rho^{\prime})^m\\
&= 1 - a^m
\end{align}

By similar reasoning, we can say:
\begin{align}
\textrm{Pr}(H_1^{-}) &= 1 - (\gamma a)^m
\end{align}

Next, we work out the probability for the event $H_2^{+}$, which which occurs iff at least two of $\tilde{y}_1^{+},...,\tilde{y}_m^{+}$ are $1$:
\begin{align}
\textrm{Pr}(H_2^{+}) &= 1 - (1 - \rho^{\prime})^m - m \rho^{\prime} (1 - \rho^{\prime})^{m-1}\\
&= 1 - a^m - m (1 - a)a^{m-1}\\
&= 1 - a^m - A
\end{align}

Similar reasoning can be used for $H_2^{-}$:
\begin{align}
\textrm{Pr}(H_2^{-}) &= 1 - (\gamma a)^m - m (1 - \gamma a)(\gamma a)^{m-1}\\
&= 1 - (\gamma a)^m - G
\end{align}

Putting these all together, we get:
\begin{align}
d^{HH} &= \textrm{Pr}(H_2^{+}) \textrm{Pr}(H_2^{-}) - \textrm{Pr}(H_1^{+}) \textrm{Pr}(H_1^{-})\\
&= (1 - a^m - A) (1 - (\gamma a)^m - G) - (1 - a^m) (1 - (\gamma a)^m)\\
&= (1 - a^m) (1 - (\gamma a)^m) - (1 - a^m) G - (1 - (\gamma a)^m) A + AG - (1 - a^m) (1 - (\gamma a)^m)\\
&= AG - (1 - a^m) G - (1 - (\gamma a)^m) A\\
d^{HL} &= \textrm{Pr}(H_2^{+}) \textrm{Pr}(L_2^{-}) - \textrm{Pr}(H_1^{+}) \textrm{Pr}(L_1^{-})\\
&= (1 - a^m - A) ((\gamma a)^m + G) - (1 - a^m) (\gamma a)^m\\
&= (1 - a^m) (\gamma a)^m + (1 - a^m) G - (\gamma a)^m A - AG - (1 - a^m) (\gamma a)^m\\
&= (1 - a^m) G - AG - (\gamma a)^m A\\
d^{LH} &= \textrm{Pr}(L_2^{+}) \textrm{Pr}(H_2^{-}) - \textrm{Pr}(L_1^{+}) \textrm{Pr}(H_1^{-})\\
&= (a^m + A) (1 - (\gamma a)^m - G) - (a^m) (1 - (\gamma a)^m)\\
&= (a^m) (1 - (\gamma a)^m) + (1 - (\gamma a)^m) A - a^m G - AG - (a^m) (1 - (\gamma a)^m)\\
&= (1 - (\gamma a)^m) A - AG - a^m G\\
d^{LL} &= \textrm{Pr}(L_2^{+}) \textrm{Pr}(L_2^{-}) - \textrm{Pr}(L_1^{+}) \textrm{Pr}(L_1^{-})\\
&= (a^m + A) ((\gamma a)^m + G) - (a^m) (\gamma a)^m\\
&= (a^m) (\gamma a)^m + a^m G + (\gamma a)^m A + AG - (a^m) (\gamma a)^m\\
& = AG + a^m G + (\gamma a)^m A
\end{align}
\end{proof}

Now that we have derived these probability differences, we can use them to work out $\textrm{Pr}(W_2) - \textrm{Pr}(W_1)$ for the four possible settings of $(\tilde{y}_0^{+}, \tilde{y}_0^{-})$. We address these case-by-case.

\paragraph{Case 1: $(\tilde{y}_0^{+}, \tilde{y}_0^{-}) = (0,0)$} We can think of this as the \textbf{``TN-vs-FN'' case}, where object $0$ is not visible in either of the ground-truth positive or ground-truth negative images. In this case, we can derive the following expression:

\begin{lemma}
\label{lemma:lemma3}
If $(\tilde{y}_0^{+}, \tilde{y}_0^{-}) = (0,0)$ then $\textrm{Pr}(W_2) - \textrm{Pr}(W_1) = \frac{1}{2} (G - A)$.
\end{lemma}

\begin{proof}
In this case, we have $\bar{r}_1^{+}, \bar{r}_2^{+}, \bar{r}_1^{-}, \bar{r}_2^{-} \in \{0, 1\}$, so we can say:
\begin{align}
\textrm{Pr}(W_1) &= \frac{1}{2} \textrm{Pr}(H_1^{+}, H_1^{-}) + \frac{1}{2} \textrm{Pr}(L_1^{+}, L_1^{-}) + \textrm{Pr}(H_1^{+}, L_1^{-})\\
&= \frac{1}{2} \Big( \textrm{Pr}(H_1^{+}) \textrm{Pr}(H_1^{-}) + \textrm{Pr}(L_1^{+}) \textrm{Pr}(L_1^{-}) + 2 \textrm{Pr}(H_1^{+}) \textrm{Pr}(L_1^{-}) \Big)\\
\textrm{Pr}(W_2) &= \frac{1}{2} \Big( \textrm{Pr}(H_2^{+}) \textrm{Pr}(H_2^{-}) + \textrm{Pr}(L_2^{+}) \textrm{Pr}(L_2^{-}) + 2 \textrm{Pr}(H_2^{+}) \textrm{Pr}(L_2^{-}) \Big)
\end{align}

Hence, we can express the win-rate difference using $d^{HH}, d^{HL}, d^{LH}, d^{LL}$ as follows:
\begin{align}
\textrm{Pr}(W_2)\! -\! \textrm{Pr}(W_1) &= \frac{1}{2} \Big(  d^{HH} + 2 d^{HL} + d^{LL}  \Big)\\
&= \frac{1}{2} \Big( AG\! -\! (1\! -\! a^m) G\! -\! (1\! -\! (\gamma a)^m) A + 2 ( (1\! -\! a^m) G\! -\! AG\! -\! (\gamma a)^m A ) + AG\! +\! a^m G\! +\! (\gamma a)^m A \Big)\\
&= \frac{1}{2} (G - A)
\end{align}
We note that this quantity is positive for sufficiently large $m$ because $\frac{G}{A} = \frac{1-\gamma a}{1 - a} \gamma^{m-1} = \frac{q^{\prime}}{\rho^{\prime}} (\frac{1 - q^{\prime}}{1 - \rho^{\prime}})^{m-1}$, which grows with $m$ because $\rho^{\prime} > q^{\prime}$ per Lemma~\ref{lemma:lemma1}.
\end{proof}

\paragraph{Case 2: $(\tilde{y}_0^{+}, \tilde{y}_0^{-}) = (1,1)$} We can think of this as the \textbf{``TP-vs-FP'' case}, where object $0$ is visible (correctly or spuriously) in both the ground-truth positive and ground-truth negative images. This leads to the same win-rate difference as the previous case.

\begin{lemma}
\label{lemma:lemma4}
If $(\tilde{y}_0^{+}, \tilde{y}_0^{-}) = (1,1)$ then $\textrm{Pr}(W_2) - \textrm{Pr}(W_1) = \frac{1}{2} (G - A)$.
\end{lemma}

\begin{proof}
In this case, we have $\bar{r}_1^{+}, \bar{r}_2^{+}, \bar{r}_1^{-}, \bar{r}_2^{-} \in \{1, 1 + \delta\}$, so we can say:
\begin{align}
\textrm{Pr}(W_1) &= \frac{1}{2} \Big( \textrm{Pr}(H_1^{+}) \textrm{Pr}(H_1^{-}) + \textrm{Pr}(L_1^{+}) \textrm{Pr}(L_1^{-}) + 2 \textrm{Pr}(H_1^{+}) \textrm{Pr}(L_1^{-}) \Big)\\
\textrm{Pr}(W_2) &= \frac{1}{2} \Big( \textrm{Pr}(H_2^{+}) \textrm{Pr}(H_2^{-}) + \textrm{Pr}(L_2^{+}) \textrm{Pr}(L_2^{-}) + 2 \textrm{Pr}(H_2^{+}) \textrm{Pr}(L_2^{-}) \Big)
\end{align}

Hence, the win-rate difference is the same as before, via the same steps as the previous lemma (Lemma~\ref{lemma:lemma3}):
\begin{align}
\textrm{Pr}(W_2) - \textrm{Pr}(W_1) &= \frac{1}{2} (G - A)
\end{align}
\end{proof}

\paragraph{Case 3: $(\tilde{y}_0^{+}, \tilde{y}_0^{-}) = (0,1)$} We can think of this as the \textbf{``FP-vs-FN'' case}, where object $0$ is occluded or obscured in the ground-truth positive image and spuriously visible in the ground-truth negative image. This is the most ``difficult'' case to rectify.

\begin{lemma}
\label{lemma:lemma5}
If $(\tilde{y}_0^{+}, \tilde{y}_0^{-}) = (0,1)$ then $\textrm{Pr}(W_2) - \textrm{Pr}(W_1) = \frac{1}{2} G \Big(1 - a^{m-1}(1 - \rho^{\prime}  + m \rho^{\prime} +  \rho^{\prime} \frac{1 - q^{\prime}}{q^{\prime}})  \Big)$.
\end{lemma}

\begin{proof}
In this case, we have $\bar{r}_1^{+}, \bar{r}_2^{+} \in \{0, 1\}$ and $\bar{r}_1^{-}, \bar{r}_2^{-} \in \{1, 1 + \delta\}$. The best we can hope for is a tie, where the positive example takes on its higher value \textit{and} the negative example takes on its lower value. Hence:
\begin{align}
\textrm{Pr}(W_1) &= \frac{1}{2} \textrm{Pr}(H_1^{+}, L_1^{-}) = \frac{1}{2} \textrm{Pr}(H_1^{+}) \textrm{Pr}(L_1^{-})\\
\textrm{Pr}(W_2) &= \frac{1}{2} \textrm{Pr}(H_2^{+}, L_2^{-}) = \frac{1}{2} \textrm{Pr}(H_2^{+}) \textrm{Pr}(L_2^{-})
\end{align}

Hence, we can express the win-rate difference as:
\begin{align}
\textrm{Pr}(W_2) - \textrm{Pr}(W_1) &= \frac{1}{2} d^{HL}\\
&= \frac{1}{2} \Big(  (1 - a^m) G - AG - (\gamma a)^m A  \Big)\\
&= \frac{1}{2} G \Big(  (1 - a^m)  - A - (\gamma a)^m \frac{A}{G}  \Big)\\
&= \frac{1}{2} G \Big(  (1 - a^m)  - A -  a^m \gamma \frac{1 - a}{1 - \gamma a}  \Big)\\
&= \frac{1}{2} G \Big(  (1 - a^m)  - m (1 - a) a^{m-1} -  a^m \gamma \frac{1 - a}{1 - \gamma a}  \Big)\\
&= \frac{1}{2} G \Big(1 - a^{m-1}(a  + m (1 - a) +  a \gamma \frac{1 - a}{1 - \gamma a})  \Big)\\
&= \frac{1}{2} G \Big(1 - a^{m-1}(1 - \rho^{\prime}  + m \rho^{\prime} +  \rho^{\prime} \frac{1 - q^{\prime}}{q^{\prime}})  \Big)
\end{align}
We once again note that this quantity is positive for sufficiently large $m$ because $a^{m-1}(1\! -\! \rho^{\prime}\!  +\! m \rho^{\prime}\! +\!  \rho^{\prime} \frac{1\! -\! q^{\prime}}{q^{\prime}}) = o(m a^m)$\\ since $a = (1 - \rho^{\prime}) < 1$ (per Lemma~\ref{lemma:lemma1}).
\end{proof}

\paragraph{Case 4: $(\tilde{y}_0^{+}, \tilde{y}_0^{-}) = (1,0)$} We can think of this as the \textbf{``TP-vs-TN'' case}, where there is no occlusion or spurious cue for object $0$ in either image. This is the ``easiest'' case to deal with, and it turns out to be the one case where a first-max is actually better than a second-max.

\begin{lemma}
\label{lemma:lemma6}
If $(\tilde{y}_0^{+}, \tilde{y}_0^{-}) = (1,0)$ then $\textrm{Pr}(W_2) - \textrm{Pr}(W_1) = \frac{1}{2} A \Big(G + a (1 - q^{\prime})^{m-1} (\frac{q^{\prime}}{\rho^{\prime}} + \frac{1 - q^{\prime}}{1 - \rho^{\prime}}) - 1 \Big)$.
\end{lemma}

\begin{proof}
In this case, we have $\bar{r}_1^{+}, \bar{r}_2^{+} \in \{1, 1+\delta\}$ and $\bar{r}_1^{-}, \bar{r}_2^{-} \in \{0, 1\}$. The worst thing that can happen is a ``tie'', in the event that the positive example gets its lower value \textit{and} the negative example gets its higher one, otherwise we get an outright ``win''. Hence:
\begin{align}
\textrm{Pr}(W_1) &= \textrm{Pr}(H_1^{+}, H_1^{-}) + \textrm{Pr}(H_1^{+}, L_1^{-}) + \frac{1}{2} \textrm{Pr}(L_1^{+}, H_1^{-}) + \textrm{Pr}(L_1^{+}, L_1^{-})\\
&= 1 - \frac{1}{2} \textrm{Pr}(L_1^{+}, H_1^{-})\\
&= 1 - \frac{1}{2} \textrm{Pr}(L_1^{+}) \textrm{Pr}(H_1^{-})\\
\textrm{Pr}(W_2) &= 1 - \frac{1}{2} \textrm{Pr}(L_2^{+}) \textrm{Pr}(H_2^{-})
\end{align}

Hence, we can express the win-rate difference as:
\begin{align}
\textrm{Pr}(W_2) - \textrm{Pr}(W_1) &= -\frac{1}{2} d^{LH}\\
&= -\frac{1}{2} \Big( (1 - (\gamma a)^m) A - AG - a^m G \Big)\\
&= \frac{1}{2} \Big( AG + a^m G - (1  - (\gamma a)^m) A \Big)\\
&= \frac{1}{2} A \Big(G + a^m \frac{G}{A} - (1  - (\gamma a)^m) \Big)\\
&= \frac{1}{2} A \Big(G + a^m \frac{1 - \gamma a}{1 - a} \gamma^{m-1} + (\gamma a)^m - 1 \Big)\\
&= \frac{1}{2} A \Big(G + a (\gamma a)^{m-1} (\frac{1 - \gamma a}{1 - a} + \gamma) - 1 \Big)\\
&= \frac{1}{2} A \Big(G + a (1 - q^{\prime})^{m-1} (\frac{q^{\prime}}{\rho^{\prime}} + \frac{1 - q^{\prime}}{1 - \rho^{\prime}}) - 1 \Big)
\end{align}
We note that in this particular case, $\textrm{Pr}(W_2) - \textrm{Pr}(W_1)$ actually becomes \textit{negative} as $m$ grows large, because both $G$ and $a (1 - q^{\prime})^{m-1} (\frac{q^{\prime}}{\rho^{\prime}} + \frac{1 - q^{\prime}}{1 - \rho^{\prime}})$ shrink exponentially with $m$, and so $\frac{1}{2} A$ is eventually multiplied by a negative number. However, we will soon see that this negative win-rate difference is outweighed by the positive ones from Lemmas~\ref{lemma:lemma3},~\ref{lemma:lemma4}, and~\ref{lemma:lemma5}, leading to an overall positive difference.
\end{proof}

We have now worked out the win-rate differences for all four possible values of $(\tilde{y}_0^{+}, \tilde{y}_0^{-})$. We are finally ready to prove our main theorem, which we restate below.\\

\restate{thm:theorem1}{If $\textrm{Pr}(\tilde{y}_0^{+} \neq y_0^{+} \bigvee \tilde{y}_0^{-} \neq y_0^{-}) > 0$, then $\textrm{Pr}(W_2) > \textrm{Pr}(W_1)$ for sufficiently large $m$.}

\begin{proof}
We start by defining a distribution over $(\tilde{y}_0^{+}, \tilde{y}_0^{-})$:
\begin{align}
\pi^{(0,0)} &:= \textrm{Pr}(\tilde{y}_0^{+} = 0,\ \tilde{y}_0^{-} = 0)\\
\pi^{(1,1)} &:= \textrm{Pr}(\tilde{y}_0^{+} = 1,\ \tilde{y}_0^{-} = 1)\\
\pi^{(0,1)} &:= \textrm{Pr}(\tilde{y}_0^{+} = 0,\ \tilde{y}_0^{-} = 1)\\
\pi^{(1,0)} &:= \textrm{Pr}(\tilde{y}_0^{+} = 1,\ \tilde{y}_0^{-} = 0)
\end{align}

We can combine the results of Lemmas~\ref{lemma:lemma3},~\ref{lemma:lemma4},~\ref{lemma:lemma5}, and~\ref{lemma:lemma6}, to get the overall win-rate difference:
\begin{align}
\label{supeq:theorem1_exact_difference}
\textrm{Pr}(W_2) - \textrm{Pr}(W_1)\quad = \qquad &\frac{1}{2} (\pi^{(0,0)} + \pi^{(1,1)}) (G - A)\\
+ &\frac{1}{2} \pi^{(0,1)}  G \Big(1 - a^{m-1}(1 - \rho^{\prime}  + m \rho^{\prime} +  \rho^{\prime} \frac{1 - q^{\prime}}{q^{\prime}})  \Big)\\
+ &\frac{1}{2} \pi^{(1,0)} A \Big(G + a (1 - q^{\prime})^{m-1} (\frac{q^{\prime}}{\rho^{\prime}} + \frac{1 - q^{\prime}}{1 - \rho^{\prime}}) - 1 \Big)
\end{align}

We can lower-bound the last term of this sum with $-A$ to get:
\begin{align}
\textrm{Pr}(W_2) - \textrm{Pr}(W_1)\quad \geq \qquad &\frac{1}{2} (\pi^{(0,0)} + \pi^{(1,1)}) (G - A)\\
+ &\frac{1}{2} \pi^{(0,1)}  G \Big(1 - a^{m-1}(1 - \rho^{\prime}  + m \rho^{\prime} +  \rho^{\prime} \frac{1 - q^{\prime}}{q^{\prime}})  \Big)\\
- &\frac{1}{2} \pi^{(1,0)} A
\end{align}

If we pick $m > 1 + \frac{\log(2) + \log(1 - \rho^{\prime}  + m \rho^{\prime} +  \rho^{\prime} \frac{1 - q^{\prime}}{q^{\prime}})}{-\log(1 - \rho^{\prime})}$ then we can lower-bound the second term to get:
\begin{align}
\textrm{Pr}(W_2) - \textrm{Pr}(W_1)\quad \geq \qquad &\frac{1}{2} (\pi^{(0,0)} + \pi^{(1,1)}) (G - A)\\
+ &\frac{1}{4} \pi^{(0,1)}  G\\
- &\frac{1}{2} \pi^{(1,0)} A
\end{align}

which simplifies to:
\begin{align}
\textrm{Pr}(W_2) - \textrm{Pr}(W_1) &\geq \frac{1}{2} \Big(  (\pi^{(0,0)} + \pi^{(1,1)} + \frac{1}{2} \pi^{(0,1)}) G - (\pi^{(0,0)} + \pi^{(1,1)} + \pi^{(1,0)}) A \Big)\\
&= \frac{1}{2} \Big(  (\pi^{(0,0)} + \pi^{(1,1)} + \frac{1}{2} \pi^{(0,1)}) G - (1 - \pi^{(0,1)}) A \Big)
\end{align}

We note that $\pi^{(0,0)} + \pi^{(1,1)} + \frac{1}{2} \pi^{(0,1)} > 0$ due to our assumption that $\textrm{Pr}(\tilde{y}_0^{+} \neq y_0^{+} \bigvee \tilde{y}_0^{-} \neq y_0^{-}) > 0$. Hence, we can do some rearrangement to get:
\begin{align}
\textrm{Pr}(W_2) - \textrm{Pr}(W_1) &\geq \frac{1}{2} (\pi^{(0,0)} + \pi^{(1,1)} + \frac{1}{2} \pi^{(0,1)}) G \Big(1 - (\frac{1 - \pi^{(0,1)}}{\pi^{(0,0)} + \pi^{(1,1)} + \frac{1}{2} \pi^{(0,1)}}) (\frac{A}{G}) \Big)\\
&= \frac{1}{2} (\pi^{(0,0)} + \pi^{(1,1)} + \frac{1}{2} \pi^{(0,1)}) G \Big(1 - (\frac{1 - \pi^{(0,1)}}{\pi^{(0,0)} + \pi^{(1,1)} + \frac{1}{2} \pi^{(0,1)}}) (\frac{1 - a}{1 - \gamma a}) \gamma^{-(m-1)})\\
&= \frac{1}{2} (\pi^{(0,0)} + \pi^{(1,1)} + \frac{1}{2} \pi^{(0,1)}) G \Big(1 - (\frac{1 - \pi^{(0,1)}}{\pi^{(0,0)} + \pi^{(1,1)} + \frac{1}{2} \pi^{(0,1)}}) (\frac{\rho^{\prime}}{q^{\prime}}) (\frac{1 - \rho^{\prime}}{1 - q^{\prime}})^{m-1})
\end{align}

We can make this expression positive by picking $m > 1 + \frac{\log(\rho^{\prime}) - \log(q^{\prime}) + \log(1 - \pi^{(0,1)}) - \log(\pi^{(0,0)} + \pi^{(1,1)} + \frac{1}{2} \pi^{(0,1)})}{\log(1 - q^{\prime}) - \log(1 - \rho^{\prime})}$.\\

Thus, the second-max will be advantageous over the first-max as long as $m$ satisfies the following lower bounds:
\begin{align}
m &> 1 + \frac{\log(2) + \log(1 - \rho^{\prime}  + m \rho^{\prime} +  \rho^{\prime} \frac{1 - q^{\prime}}{q^{\prime}})}{-\log(1 - \rho^{\prime})}\\
m &> 1 + \frac{\log(\rho^{\prime}) - \log(q^{\prime}) + \log(1 - \pi^{(0,1)}) - \log(\pi^{(0,0)} + \pi^{(1,1)} + \frac{1}{2} \pi^{(0,1)})}{\log(1 - q^{\prime}) - \log(1 - \rho^{\prime})}
\end{align}

In fact, in the special case where $\pi^{(1,0)} = 0$, these bounds are ``tight'' in the sense that an $m$ that violates both bounds will lead to the first-max being advantageous over the second-max. Theorem~\ref{thm:theorem2} gives further insight into this dependency on $m$.

\end{proof}

\subsection{Proof of Theorem 2}
\label{sec:theorem2_proof}
\begin{proof}
We prove Theorem \ref{thm:theorem2} by example. We give two example settings that fulfill the requirements of Theorem \ref{thm:theorem1} and for which $\textrm{Pr}(W_2) < \textrm{Pr}(W_1)$ for sufficiently small $m$. Furthermore, these two example settings differ only by $\nu$ and have different values of $m$ at which the inequality reverses, and so we know that this reversal point depends on $\nu$. Some further examples show that the reversal point also depends on other setting variables such as $\rho, q, \pi^{(0,0)}, \pi^{(1,1)},\pi^{(0,1)}, \pi^{(1,0)}$, but we limit our analysis to $\nu$ for the sake of brevity.

\begin{figure}[t]
  \centering
    \includegraphics[width=1\linewidth]{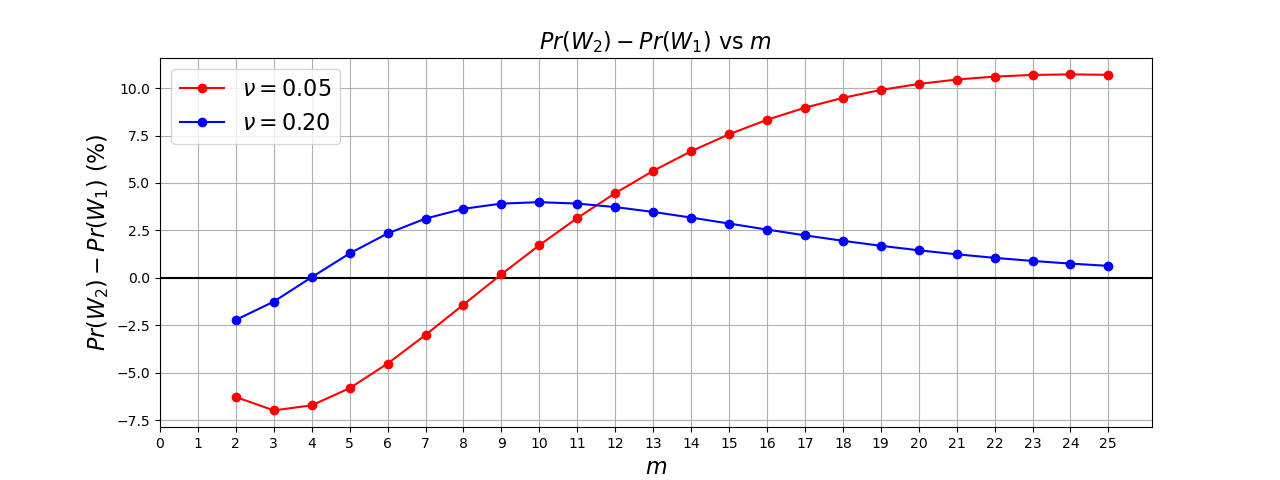}
    \caption{$\textrm{Pr}(W_2) - \textrm{Pr}(W_1)$ via Eq.~\eqref{supeq:theorem1_exact_difference} for different values of $m$, proving by example that first-max can be advantageous for sufficiently small $m$, and that the point of advantage depends on label-flip probability $\nu$.}
    \label{fig:theorem2_example}
\end{figure}

Our two example settings share $\rho = 0.15,\ q = 0.01,\ \pi^{(1,0)} = 0.55^2,\ \pi^{(0,1)} = (1 - 0.55)^2$, and $\pi^{(0,0)} = \pi^{(1,1)} = 0.55 \cdot (1 - 0.55)$. They differ only in their values of $\nu$ which are $0.05$ and $0.2$. We evaluate $\textrm{Pr}(W_2) - \textrm{Pr}(W_1)$ via Eq.~\eqref{supeq:theorem1_exact_difference} for multiple values of $m$ under these settings and plot the resulting probability differences in Fig.~\ref{fig:theorem2_example}. We see that the claims in Theorem \ref{thm:theorem2} follow from these examples.
\end{proof}

\end{document}